\documentclass[journal]{IEEEtran}

\usepackage{ifpdf}  % MISC UTILITY PACKAGES
\usepackage{cite}  % CITATION PACKAGES
\ifCLASSINFOpdf
  \usepackage[pdftex]{graphicx}
\else
  \usepackage[dvips]{graphicx}
\fi

\usepackage{amsmath}  % MATH PACKAGES
\usepackage{newtxtext}
\usepackage{newtxmath}
\usepackage{placeins}
\usepackage{algorithmic}  % SPECIALIZED LIST PACKAGES
\usepackage{array}  % ALIGNMENT PACKAGES

\ifCLASSOPTIONcompsoc
 \usepackage[caption=false,font=normalsize,labelfont=sf,textfont=sf]{subfig}
\else
 \usepackage[caption=false,font=footnotesize]{subfig}
\fi

\usepackage[utf8]{inputenc} % allow utf-8 input
\usepackage[T1]{fontenc}    % use 8-bit T1 fonts
\usepackage{hyperref}       % hyperlinks
\usepackage{url}            % simple URL typesetting
\usepackage{booktabs}       % professional-quality tables
\usepackage{amsfonts}       % blackboard math symbols
\usepackage{nicefrac}       % compact symbols for 1/2
\usepackage{microtype}      % microtypography
\usepackage{enumitem}       % Adjusting line spacing of itemize
\usepackage{lipsum}         % random paragraph

\usepackage{mathrsfs}       % calligraphic letters
\usepackage{bbm}            % double line letter

\usepackage{multirow}       % table format
\usepackage{fancyhdr}       % header
\usepackage{algorithm}

\begin{document}
\title{Structure-Guided Spatiotemporal Attention Graph Neural Network for Traffic Flow Prediction}

\author{Xuanmian~He,
        Can~Li,
        Wanjing~Ma
\thanks{This research is supported by [52402407, the National Natural Science Foundation of China]. \textit{(Corresponding author: Can Li.)}}
\thanks{Xuanmian He is with the Department of Civil and Environmental Engineering, University of California, Berkeley, United States. Can Li is with the Key Laboratory of Road and Traffic Engineering of the Ministry of Education, Tongji University, Shanghai 201804, China (e-mail: \href{mailto:xuanmianh@berkeley.edu}{xuanmianh@berkeley.edu}, \href{mailto:lican@tongji.edu.cn}{lican@tongji.edu.cn}, \href{mailto:mawanjing@tongji.edu.cn}{mawanjing@tongji.edu.cn}}}

\maketitle
\markboth{}%
{He \MakeLowercase{\textit{et al.}}: Structure-Guided Spatiotemporal Attention Graph Neural Network for Traffic Flow Prediction}

\begin{abstract}
Deep spatiotemporal models integrating graph convolutions and attention mechanisms have demonstrated excellent performance in network-level traffic flow prediction, owing to their exceptional ability to capture complex spatiotemporal dependencies. Despite their predictive success, deployment of such models in safety-critical urban systems remains constrained by their inherent lack of transparency. Existing post-hoc diagnostic methods often struggle with spurious correlations and fail to unveil the intrinsic decision-making mechanisms governing traffic dynamics, resulting in suboptimal interpretability and limited operational trustworthiness. To address these challenges, this paper proposes the Structure-Guided Spatiotemporal Attention Graph Neural Network (SGSAN). Departing from traditional architectures that rely on unconstrained adaptive graphs, SGSAN explicitly learns a static Directed Dependency Graph (DDG) to identify the invariant macroscopic propagation paths of traffic states. We further introduce an InfoNCE-based soft-coupling mechanism that anchors the model's dynamic spatiotemporal attention to this structural prior, offering a mechanistic account of the model's decision-making process while ensuring robust forecasting by aligning attention-based reasoning with identified macroscopic dependencies and preventing over-reliance on ephemeral local noise. Furthermore, a decoupled two-stage optimization framework is developed to resolve the fundamental conflict between structural discovery and predictive error minimization. Extensive experiments on multiple real-world datasets demonstrate that SGSAN achieves state-of-the-art predictive accuracy while providing built-in interpretability that organically aligns with the physical logic of traffic networks. Our findings highlight that explicitly modeling directed structural dependencies can transform opaque spatiotemporal correlations into transparent, mechanistic insights without sacrificing predictive performance.
\end{abstract}


\IEEEpeerreviewmaketitle

\section{Introduction}

% background
Accurate and reliable spatiotemporal traffic flow prediction plays a vital role in Intelligent Transportation Systems (ITS), providing the analytical foundation for proactive traffic control, efficient congestion mitigation, and optimal route planning \cite{Vlahogianni}. With the pervasive deployment of road network sensors, deep learning models have achieved high prediction accuracy by capturing complex features across large-scale networks \cite{Yin, Lv, Tian}. However, the transition of these models from laboratory benchmarks to safety-critical urban systems is hindered by their "black-box" nature \cite{Wang}. 

In practical traffic engineering, achieving reliable results requires anchored predictions that supplement high accuracy with structural interpretability. The opacity of complex architectures often leads to the capture of spurious relations, where models may achieve low training error by over-fitting to ephemeral local noise rather than modeling the stable, underlying network structure \cite{ZhangTino}.

% interpretable DNN and Attention mechanism
To address this lack of transparency, an increasing number of studies have focused on developing interpretable frameworks. Existing efforts primarily fall into three categories: post-hoc diagnostics (such as LIME and SHAP \cite{Ribeiro, lundberg2017}) , gradient-based attribution methods \cite{bach2015pixel}, and attention-based mechanisms. While post-hoc methods can generate plausible rationalizations for predictions, they operate as external estimators that merely approximate model behavior without reflecting the actual internal decision-making mechanisms. Alternatively, although attention mechanisms provide a degree of transparency by highlighting important features \cite{Do}, they are fundamentally correlation-driven \cite{wiegreffe2019}. In complex traffic environments, these unconstrained attention weights greedily aggregate a broad range of statistical signals to minimize immediate prediction error. This often includes spurious correlations, e.g., a sudden localized rainstorm might cause simultaneous congestion in two geographically disconnected zones. A purely data-driven predictive attention module will heavily rely on the strong statistical correlation despite the absence of a direct physical influence between zones. Consequently, the reasoning process remains loosely constrained and highly susceptible to noise, failing to provide the structural consistency required for engineering trust.

% Causality to DDG
While true interventionist causal inference is unachievable purely from observational traffic data without accounting for hidden confounders, the discovery techniques can be repurposed to learn a macroscopic structural prior \cite{Janzing}. In the context of traffic networks, dominant propagation paths are fundamentally dictated by the time-invariant macroscopic spatial topology. Incorporating such structural discovery into spatiotemporal learning enables the model to distinguish a stable global backbone from incidental local fluctuations or spurious correlations \cite{murdoch2019}. By anchoring dynamic reasoning to this learned backbone, the model moves beyond simple curve-fitting, ensuring that its predictive logic remains consistent with the long-term dependency patterns of the road topology.

Despite its potential, integrating structural priors with dynamic attention mechanisms faces a primary challenge: distilling a stable structural backbone from highly volatile traffic patterns without sacrificing the model's sensitivity to real-time variations. This fundamental conflict stems from two main issues. First, directly optimizing a graph structure alongside a deep predictive model often leads to gradient interference, where short-term prediction errors can distort the long-term dependency graph, resulting in an unstable topology \cite{franceschi2019}. Second, imposing rigid structural constraints can over-regularize the model, hindering its ability to respond to localized, non-topological events (e.g., accidents) \cite{jiang2022graph}.

To address this, we propose learning a static Directed Dependency Graph (DDG) to capture time-invariant global spatial dependencies, while retaining a spatiotemporal attention mechanism to adapt to dynamic local inputs. We adopt a two-stage training framework that explicitly decouples structure learning (Stage I) from spatiotemporal correlation extraction (Stage II). This separation stabilizes the graph generation process and prevents disruption from local, short-term noise \cite{jin2020graph}. Furthermore, instead of imposing overly rigid constraints, we utilize InfoNCE contrastive learning as a form of soft regularization \cite{Oord, zhang2023}. By maximizing the mutual information between the multi-scale representations and the structural prior, this soft-coupling guides the attention mechanism to anchor onto the stable DDG while preserving the flexibility necessary to capture real-time, non-topological local variations.

The main contributions of this study are summarized as follows:

\begin{itemize}
\item We propose a structure-guided spatiotemporal architecture to overcome the reliance on unconstrained observational correlations and achieve a paradigm shift towards discovering time-invariant structural priors. Specifically, by learning a Directed Dependency Graph (DDG), our approach explicitly guides the attention-based prediction process to follow macroscopic propagation paths, mitigating the impact of spurious correlations and local noise.

\item We design the Structure-Guided Spatiotemporal Attention Graph Neural Network (SGSAN) to achieve structural stability and temporal flexibility while improving training and inference efficiency. Specifically, we employ a decoupled two-stage training strategy alongside a soft-coupling mechanism that anchors dynamic attention to the static dependency graph, which has been shown to benefit convergence and prediction performance.

\item To evaluate the trustworthiness of the proposed framework, we introduce a new set of metrics that quantify the mechanistic alignment between model reasoning and physical network topology. Results demonstrate the alignment between learned dependency priors, attention maps, and physical road networks. Experiments on several real-world datasets demonstrate that SGSAN achieves state-of-the-art predictive accuracy while providing built-in interpretability.
\end{itemize}
 
\section{Related Works}
\label{sec:related_works}

\subsection{Interpretable Prediction}
Previous studies on interpretable prediction primarily fall into two categories \cite{Arrieta}. 

\begin{enumerate}
\item Inherently interpretable models, such as linear regression and ARIMA. Despite the clear physical implications, they often struggle to extract high-dimensional nonlinear features and exhibit lower accuracy in complex traffic networks. 

\item Approaches focusing on the explainability of ``black-box'' models \cite{Liang}, including post-hoc diagnostic methods, physics-informed methods, attention-based methods, and causal analysis. 
\end{enumerate}

Post-hoc diagnostic methods are frequently employed to illustrate prediction results. Cui \emph{et al.} \cite{Cui} visualized model representation weights in graph convolutional recurrent neural networks, analyzed the spatial features of traffic states, and provided explanations for specific results. Zhai \emph{et al.} \cite{Zhai} introduced a post-hoc explainability measurement method for traffic prediction issues based on sensitivity and partial dependence graph correlation tests. Medrano \emph{et al.} \cite{Medrano} employed SHAP, which establishes a dense layer after the traffic state prediction model to offer diagnostics for prediction results. Ribeiro \emph{et al.} \cite{Ribeiro} compared and analyzed the global and local explanatory capabilities of diagnostic methods such as SHAP and LIME for travel time prediction. Although these methods are model-agnostic, their reliance on local approximations limits understanding of the neural network's prediction logic and thus constrains the reliability of their explainability.

% Physics-Informed Neural Network (PINN) incorporates interpretability by introducing macroscopic traffic flow models to data-driven methods. Shi \emph{et al.} \cite{Shi} used multi-dimensional data—such as traffic volume, density, and travel time—along with the LWR model, employing physical information as soft constraints to enhance their prediction model. Zhang \emph{et al.} \cite{ZhangMao} utilized data from detectors and floating vehicles to estimate density and applied LWR constraints. However, these approaches require extensive domain knowledge and high-quality data, which can lead to systematic errors due to necessary physical assumptions that may not hold in complex real-world scenarios.

The attention mechanism has been validated to identify the most important data features by allocating weights and considering contextual relevance, which enhances prediction accuracy \cite{Niu}. It is particularly effective for modeling long-term dependencies in sequences. Furthermore, the attention weights contribute to a better understanding of the model, thereby increasing its interpretability. In the context of traffic flow prediction, these weights indicate the significance of various traffic flow features on the prediction outcomes within a specific traffic network structure \cite{Lan, Bai, Ali}. Li \emph{et al.} \cite{Li} introduced a dynamic spatiotemporal attention mechanism for traffic flow prediction, integrating it with classical traffic flow theory to analyze the speed of traffic wave propagation and to elucidate the correlations in traffic states. However, the explanations derived from the attention mechanism are often limited to specific samples, as the attention weights are closely tied to the features of the input data. Additionally, the mechanism primarily focuses on correlations, limiting its capacity to interpret the underlying causal mechanism of the predictions.

\subsection{From Causal Analysis to Structural Discovery}
Traditional neural networks can effectively identify correlations in data samples, but these correlations often lack interpretability and are susceptible to spurious statistical associations. In contrast, learning causal or structural relationships offers insights more aligned with the intrinsic mechanisms governing the prediction process. This distinction has motivated studies on causal and structure-aware modeling for traffic forecasting, which can be organized into two lines of work: causal discovery methods and causal-constrained spatiotemporal models.

\paragraph{Pure Causal Discovery Methods} 
Early efforts to introduce causal reasoning into traffic modeling drew on classical techniques from the causal inference literature. Zhang \emph{et al.}~\cite{ZhangZheng} conducted a nonlinear Granger causality analysis using a deep learning model for traffic speed prediction, providing a principled criterion for selecting informative features within the network and establishing that directional temporal dependencies yield more stable representations than symmetric correlations. Concurrently, Yu \emph{et al.}~\cite{Yu} proposed DAG-GNN, which integrates graph neural networks with variational autoencoders to learn implicit causal relationships as a continuous directed acyclic graph optimization problem, offering a scalable alternative to combinatorial structure search. Building on this, Tygesen \emph{et al.}~\cite{Tygesen} utilized inference graphs to reveal the model's focus areas and to explain how spatial correlations are leveraged for traffic prediction, demonstrating that structural graph representations can meaningfully improve the interpretability of deep learning models.

\paragraph{Causal-Constrained Spatiotemporal Models}
Motivated by the limitations of purely data-driven attention mechanisms, recent works have incorporated explicit causal structures to mitigate spurious correlations in complex traffic networks. He \emph{et al.} \cite{he2023} introduced STGC-GNNs, which detect a Spatial-Temporal Granger Causality graph to replace static distance-based dependencies, demonstrating that data-driven causality provides a more stable representation of underlying traffic flow dynamics. Zhao \emph{et al.}~\cite{Zhao2023} proposed a Causal Conditional Hidden Markov Model (CCHMM) that disentangles the physical concepts affecting multi-modal traffic generation through a structural causal model, effectively isolating causal representations from observational noise. Yang \emph{et al.}~\cite{Yang2024} further proposed the Principal Spatio-Temporal Causal Graph Convolutional Network (PSTCGCN), which employs causal convolutions and semi-principal graph embeddings to capture spatio-temporal causal dependencies in a unified framework. More recently, Zhu \emph{et al.}~\cite{zhu2025} reinforced this paradigm by integrating causal inference with graph convolutional networks to filter spurious edges and enhance model robustness against noisy observations.

Despite their contributions, existing methods either pursue strict causal identification, which is unachievable from observational traffic data alone without handling confounders and interventional experiments~\cite{Pearl}, or focus on isolated local causal features without capturing the global, time-invariant propagation topology in the network. Rather than claiming strict causal identification, SGSAN reframes the objective as structural dependency discovery. Different from diffusion-based graph formulations that propagate traffic states along pre-defined directional road structures~\cite{zhang2021traffic}, we apply the continuous optimization techniques of causal discovery to learn a DDG that reflects the dominant propagation topology of the network without requiring interventional data.

Critically, existing methods have not resolved the fundamental tension between static global structural knowledge and the dynamic, real-time flexibility required for accurate short-term prediction. While they have attempted to incorporate causal graphs into learning frameworks, they predominantly employ rigid fusion strategies. Such hard fusion impedes the mechanistic clarity of the causal structure and sacrifices the dynamic flexibility required to capture non-topological traffic anomalies as well. To bridge this gap, SGSAN proposes a hybrid paradigm. Structural discovery is employed to generate a global dependency graph and the spatiotemporal attention is used to capture dynamic, local fluctuations. By softly coupling these two mechanisms, SGSAN effectively mitigates the influence of spurious correlations, yielding predictions that are both highly accurate and structurally interpretable.

\section{Problem Formulation}
\label{sec:problem}

A physical traffic network is mathematically defined as a graph $\mathcal{G} = (\mathcal{V}, \mathcal{E})$, where $\mathcal{V}$ is the set of $V$ detector nodes ($|\mathcal{V}| = V$) and $\mathcal{E}$ is the set of edges representing the physical road connections. The underlying topological structure of this network is represented by the physical adjacency matrix $\mathbf{A} \in \mathbb{R}^{V \times V}$. 

Let $x^{i}_{t} \in \mathbb{R}$ denote the traffic state (e.g., traffic flow) of node $i$ at time step $t$. The collective traffic state of all nodes at time $t$ is denoted as $\mathbf{X}_t=[x^{1}_{t}, x^{2}_{t}, \cdots,x^{V}_{t}] (\mathbf{X}_t \in \mathbb{R}^{V})$. The spatiotemporal traffic flow prediction problem aims to learn a mapping function $f(\cdot)$ that leverages a historical sequence of traffic states over $T$ time steps, $\mathbf{X}_{t-T+1:t} \in \mathbb{R}^{V \times T}$, alongside the structural network information, to forecast the future traffic states over the next $H$ time steps, $\mathbf{X}_{t+1:t+H} \in \mathbb{R}^{V \times H}$. 

This prediction task can be formally expressed as:
\begin{equation}
    \label{eq:formulation}
    \hat{\mathbf{X}}_{t+1:t+H} = f(\mathbf{X}_{t-T+1:t}, \mathbf{A}) 
\end{equation}

In our proposed SGSAN framework, the function $f(\cdot)$ operates by explicitly learning a DDG, denoted as $G \in \mathbb{R}^{V \times V}$, which serves as a structural prior to guide a dynamic spatiotemporal attention mechanism. A comprehensive summary of the key notations used throughout this paper is provided in Table~\ref{tab:notation}.

\begin{table}[htbp]
  \centering
  \caption{Summary of Key Notations}
    \begin{tabular}{cl}
    \toprule
    \textbf{Notation} & \textbf{Definition} \\
    \midrule
    $V$ & Number of detector nodes in the network \\
    $T$ & Length of the historical input sequence \\
    $H$ & Length of the prediction horizon \\
    $d$ & Dimension of the hidden representation layers \\
    $N$ & Number of samples in a training batch \\
    $\mathbf{X}_{t}$ & Traffic states of all nodes at time step $t$ \\
    $\mathbf{A}$ & Physical adjacency matrix of the road network ($\mathbf{A} \in \mathbb{R}^{V \times V}$) \\
    $G^*$ & Adjacency matrix of the learned DDG ($G^* \in \mathbb{R}^{V \times V}$) \\
    $G^a$ & Dynamic spatial attention weight matrix ($G^a \in \mathbb{R}^{V \times V}$) \\
    $h^n$ & Initial node embedding ($h^n \in \mathbb{R}^{V \times d}$) \\
    $h^t$ & Temporal attention representation ($h^t \in \mathbb{R}^{V \times d}$) \\
    $h^{str}$ & Structural representation derived from DDG ($h^{str} \in \mathbb{R}^{V \times d}$) \\
    $h^{att}$ & Spatiotemporal attention representation ($h^{att} \in \mathbb{R}^{V \times d}$) \\
    \bottomrule
    \end{tabular}%
  \label{tab:notation}%
\end{table}

\section{Methodology}
\label{sec:methodology}

\begin{figure*}
    \centering
    \includegraphics[width=0.9\linewidth]{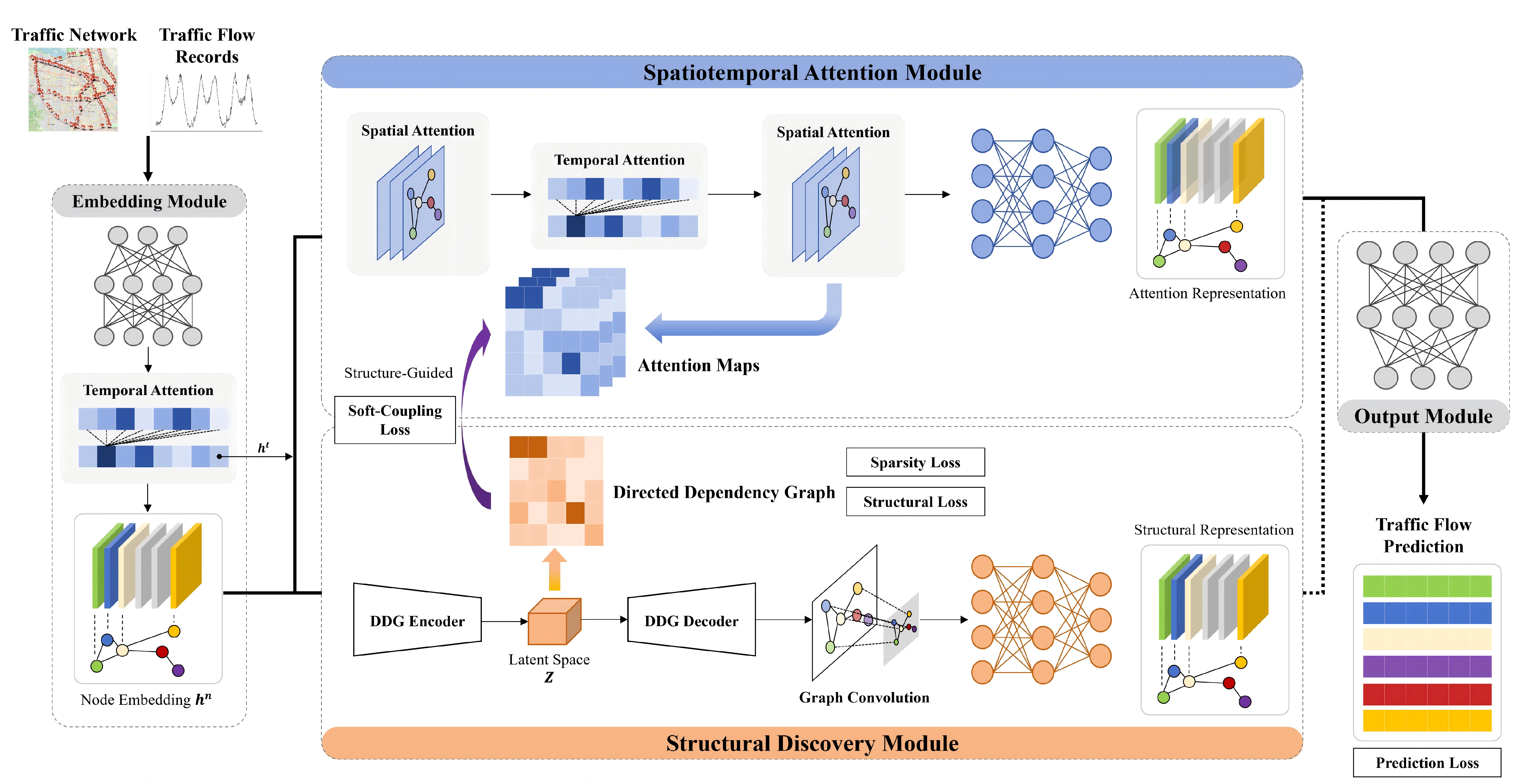}
    \caption{Framework of Structure-Guided Spatiotemporal Attention Graph Neural Network (SGSAN)}
    \label{fig:framework}
\end{figure*}

In order to effectively fuse static structural knowledge with dynamic real-time flexibility, we propose the Structure-Guided Spatiotemporal Attention Graph Neural Network (SGSAN), shown in Figure~\ref{fig:framework}. It is a four-module architecture designed to bridge structural discovery with spatiotemporal modeling. First, the embedding module employs a temporal attention sub-module to extract time-series features and project the historical states into initial node embeddings. Operating in parallel, the spatiotemporal attention module and structural discovery module are designed to extract spatiotemporal features at distinct scales. The structural discovery module learns the DDG $G^*$ to capture stable and time-invariant dependencies among the nodes, and to identify the dominant dependency backbone, as detailed in Section \ref{subsec:structural_discovery}. As detailed in Section \ref{subsec:attention}, the spatiotemporal attention module stacks two spatial attention layers and one temporal attention layer to generate the dynamic hidden state $h^{att}$. Specifically, $G^a$ denotes the weight map derived from the final spatial attention layer, encapsulating the data-driven local correlations. 

The decoder's output is then convolved to produce structural representations $h^{str}$ that capture these global topological relationships. An inherent tension exists between these two parallel tracks, \emph{i.e.}, the structural discovery module must filter out short-term spurious correlations to distill the high-level global topology, whereas the spatiotemporal attention module greedily captures localized temporal variations to maximize short-term prediction accuracy. To reconcile these divergent objectives and fuse the multi-scale information, we introduce a two-stage training framework equipped with InfoNCE contrastive learning. As detailed in Section \ref{sub_sec:loss}, this soft-coupling mechanism resolves optimization conflicts by ensuring the dynamic attention is mechanically guided by the static structural prior. Finally, the output module utilizes fully connected layers to generate the ultimate prediction. Depending on the specific phase of our two-stage framework, this prediction head dynamically routes the representations from either the spatiotemporal module or the structural discovery module to yield the final output. 

\subsection{Structural Discovery Module}
\label{subsec:structural_discovery}
The primary objective of the structural discovery module is to construct a directed dependency graph and generate a hidden structural representation ($h^{str}$), enabling Structure-Guided attention. This module primarily consists of an encoder-decoder structure that extracts topological information from node embeddings, enhanced by a graph convolution layer to further capture spatial relationships within the network. Instead of processing traffic flow directly, the module utilizes node embeddings that are encoded through a temporal attention layer. Each node embedding encapsulates the temporal dimension in the form of sequence information, allowing its shape to remain independent of the sequence length. The main aim of learning the dependency graph is to provide the attention module with a stable structural prior. Mathematically, the dependency graph is optimized as a directed acyclic graph (DAG) \cite{Pearl}. Inspired by \cite{Yu}, we design a structural discovery module that employs the DAG-GNN model, which is derived from structural equation models (SEM) and follows the architecture of variational autoencoders (VAE), consisting of an encoder and a decoder. This module represents structural relationships through generative graphical models, transforming the creation of the dependency graph into a continuous optimization problem. A linear SEM can be expressed as follows:

\begin{equation}
        h \leftarrow B^{T}h + Z
\end{equation}
where $B \in \mathbb{R}^{V \times V}$ represents the parameterized adjacency matrix of the graph. $h$ is the input, and $Z$ is a noise matrix. Therefore, the mapping from the noise matrix to the original input is denoted as:

\begin{equation}
        h = (\mathbf{I} - B^{T})^{-1} Z
\end{equation}
where $\mathbf{I}$ is the identity matrix. Based on this, the linear SEM can be transformed into a VAE structure:

\begin{equation}
    \label{eq:cas_vae}
    \begin{cases}
    Z = MLP[(\mathbf{I}-G^T) \cdot MLP (h^n)] \\
    \hat{h}^{str} = MLP[(\mathbf{I}-G^T)^{-1} \cdot MLP (Z)]
    \end{cases}
\end{equation}

In this framework, transformations are applied through multi-layer perceptrons ($MLP$). The encoder utilizes the node embedding $h^n$, while the decoder generates the intermediate output $\hat{h}^{str}$. The matrix $G$ represents the normalized dependency graph adjacency matrix that needs to be learned.

Following the VAE structure, a graph convolution layer is implemented to produce the final structural representation. Graph convolution extracts spatial features by aggregating information from neighboring nodes. The mathematical formulation is as follows:

\begin{equation}
        h^{str} = ReLU(\tilde{D}^{-\frac{1}{2}} \tilde{A} \tilde{D}^{-\frac{1}{2}} \hat{h}^{str} W)
\end{equation}

where $h^{str}$ denotes the structural representation, $\tilde{A}$ represents the normalized physical adjacency matrix after adding self-loops, and $\tilde{D}$ denotes the degree matrix of $\tilde{A}$. $W$ denotes the learnable coefficient matrix.

The initialization of the structural adjacency matrix $B$ aims to improve the expression of nonlinear features. Starting from a parameterized matrix $B_0$, the $sinh(\cdot)$ function is introduced to amplify the significance of strong dependencies, modulated by a temperature coefficient $\tau$. To obtain the final adjacency matrix $G$ used in Equation (\ref{eq:cas_vae}), we apply a Softmax normalization, thereby incorporating the weights of directed dependencies into the model training process. The Softmax function assigns the origin of influence for each node relative to others in the network:
\begin{equation}
    G = Softmax(sinh(\tau\cdot B_0))
\end{equation}

\subsection{Spatiotemporal Attention Module}
\label{subsec:attention}
To effectively model temporal dynamics in sequential traffic flow states, the temporal attention sub-module utilizes a multi-head self-attention mechanism, comprising keys ($K$), queries ($Q$), and values ($V$). They are generated through linear mapping of the input $h^t$, reflecting the underlying high-dimensional characteristics of the data. The structural equation of the attention module can be represented as follows:

\begin{equation}
        \alpha_{ij} = \text{Softmax}(\frac{Q_i \cdot {K_j}^T}{\sqrt{D}})
\end{equation}

\begin{equation}
        head = Att(Q_i, K, V) = \sum_{j}\alpha_{ij}V_j
\end{equation}
where $\alpha_{ij}$ represents the normalized attention weight coefficient. $Att(\cdot)$ is the weighted sum of coefficients multiplied by the value vector, incorporating temporal attention weights into the prediction process. Assuming $m$ attention heads, the outputs are concatenated along their respective dimensions and undergo a linear mapping to produce the temporal attention-based representation vector $h^t$.

To capture the complex spatial dependencies in traffic networks, we adopt a spatial attention module based on the graph attention network (GAT) \cite{velickovic}. For any node in the graph, the attention score $e_{ij}$ between its temporal attention representation $h^{t}_{i}$ and $h^{t}_{j}$ is calculated as:
\begin{equation}
        e_{ij} = \text{Softmax}_j[LeakyReLU([W_1 h^{t}_{i} \parallel W_2 h^{t}_{j}] \cdot W_3)]
\end{equation}

where $e_{ij}$ denotes the attention scores, and the symbol $\parallel$ represents the vector concatenation operation. $W_1$, $W_2$, and $W_3$ represent learnable parameter matrices. Finally, these scores are normalized with the Softmax function to produce the attention weight matrix $G^a = [e_{ij}]$. Each element in $G^a$ indicates the dynamic spatial correlation between node $i$ and node $j$. We capture the weight matrix from the last spatial attention sub-module, denoted as the "attention map", which is subsequently aligned with the structural representation $h^{str}$ to generate the final spatiotemporal representation $h^{att}$.

\subsection{Loss Function}
\label{sub_sec:loss}

To resolve the inherent optimization conflict between discovering stable structural dependencies and minimizing dynamic prediction errors, we design a two-stage training framework. The overall objective function is composed of four components: prediction loss ($\mathcal{L}_{A}$), dependency structural loss ($\mathcal{L}_{D}$), sparsity regularizer ($\mathcal{L}_{S}$), and InfoNCE soft-coupling loss ($\mathcal{L}_{I}$).

\textbf{(1) Prediction Loss ($\mathcal{L}_{A}$):} We employ Mean Squared Error (MSE) to optimize the predictive accuracy of the model:
\begin{equation}
    \label{eq:loss_a}
    \mathcal{L}_{A} = \frac{1}{N}\sum_{i=1}^{N}(\hat{y}_{t}^{i} - y_{t}^{i})^{2}
\end{equation}
where $\hat{y}_{t}^{i}$ denotes the predicted traffic state, $y_{t}^{i}$ is the ground truth, and $N$ is the number of samples in the batch.

\textbf{(2) Dependency Structural Loss ($\mathcal{L}_{D}$):} To ensure the learned dependency graph forms a valid directed acyclic structure, we enforce a continuous NOTEARS trace exponential penalty \cite{Zheng}:
\begin{equation}
    \label{eq:loss_d}
    \mathcal{L}_{D} = c(V) \cdot [\text{tr}(e^{G \odot G}) - V]
\end{equation}
where $G \in \mathbb{R}^{V \times V}$ is the adjacency matrix of the dependency graph, $V$ is the number of nodes in the physical network, $\text{tr}(\cdot)$ represents the trace of a matrix, $\odot$ denotes the Hadamard product, and $e^{(\cdot)}$ denotes the matrix exponential. Based on the properties of the matrix exponential, $\text{tr}(e^{G \odot G}) = V$ holds strictly true if and only if $G$ is an acyclic structure. The term $c(V)$ is a normalization coefficient set to $1 \times 10^{-(V // 10)}$ that can easily stabilize gradients across different graph scales.

\textbf{(3) Sparsity Regularizer ($\mathcal{L}_{S}$):} To prevent over-smoothing in spatial aggregators and eliminate redundant edges, an $L_1$ regularization term is applied:
\begin{equation}
    \label{eq:loss_s}
    \mathcal{L}_{S} = \| G \|_{1} = \sum_{i=1}^{V}\sum_{j=1}^{V}| G_{ij} | 
\end{equation}
This strict structural sparsity serves as a crucial inductive bias, aiding the model in identifying the dominant dependency backbone while reducing the influence of spurious correlations.

\textbf{(4) InfoNCE Soft-Coupling Loss ($\mathcal{L}_{I}$):} Instead of directly enforcing similarity via KL divergence—which often leads to rigid structural constraints and eliminates temporal flexibility—we utilize contrastive learning to enhance the spatiotemporal attention mechanism \cite{Oord}:
\begin{equation}
    \label{eq:loss_i}
    \mathcal{L}_{I} = -\frac{1}{N} \sum_{i=1}^{N}\log \frac{\exp(\text{sim}(h_{i}^{str}, h_{i}^{att})/\tau_0)}{\sum_{j=1}^{N}\exp(\text{sim}(h_{i}^{str}, h_{j}^{att})/\tau_0)}
\end{equation}
where $h_{i}^{str}$ denotes the structural representation derived from the dependency graph, and $h_{i}^{att}$ denotes the dynamic spatiotemporal attention representation. $\text{sim}(\cdot)$ is the cosine similarity function, and $\tau_0$ is the temperature for InfoNCE loss. This coupling mechanism operates as a soft alignment strategy, effectively anchoring the attention to the dependency skeleton while preserving the necessary representational freedom for the attention mechanism to capture real-time local dependencies.

\subsection{Two-Stage Training Strategy}

Training the directed dependency structure and the attention mechanism simultaneously forces them to optimize fundamentally different objectives concurrently, leading to mutual interference and degraded performance. To address this, we introduce an explicitly decoupled two-stage learning process, outlined in Algorithm \ref{alg:framework}.

\textbf{Stage I (DDG Generation):} The main objective is to learn a global dependency graph that incorporates time-invariant macroscopic topology. The dynamic attention module is excluded, and the structural discovery module is optimized using the following joint loss:
\begin{equation}
    \label{eq:loss_1}
    \mathcal{L}_{\text{Stage-I}} = \mathcal{L}_{A} + \gamma_{D} \cdot \mathcal{L}_{D} + \gamma_{S} \cdot \mathcal{L}_{S}
\end{equation}

\textbf{Stage II (Soft Alignment):} The structural discovery module is frozen, providing the learned dependency graph ($G^*$) as a stable structural prior. The spatiotemporal attention module is then activated and optimized. The core objective in this stage is to softly couple the multi-scale representations, mitigating spurious correlations while preserving the attention mechanism's flexibility to adapt to real-time inputs. The total loss function is defined as:
\begin{equation}
    \label{eq:loss_2}
    \mathcal{L}_{\text{Stage-II}} = \mathcal{L}_{A} + \gamma_{I} \cdot \mathcal{L}_{I}
\end{equation}
where $\gamma_{D}$, $\gamma_{S}$, and $\gamma_{I}$ are balancing coefficients determined empirically to achieve an optimal balance between directed dependency graph convergence and contrastive learning.

\begin{algorithm}[htbp]
    \caption{Two-Stage Training Framework of SGSAN}
    \label{alg:framework}
    \begin{algorithmic}[1]
        \REQUIRE Input sequences $\mathbf{X}_{t-T+1:t}$, Adjacency matrix $\mathbf{A}$, Balancing hyperparameters $\gamma_{D}, \gamma_{S}, \gamma_{I}$
        \ENSURE Predicted traffic states $\hat{\mathbf{X}}_{t+1:t+H}$
        \STATE Initialize parameters for Temporal Module $\mathcal{M}_t$, Spatial Attention Module $\mathcal{M}_s$, Structural Discovery Module $\mathcal{M}_{str}$, and Output Module $f_{out}$.
        
        \STATE \hrulefill
        \STATE \textbf{Stage I: Generate the Directed Dependency Graph (DDG)}
        \WHILE{Stage I not converged}
            \STATE $h^t = \mathcal{M}_t(\mathbf{X}_{t-T+1:t})$
            \STATE $h^{str}, G = \mathcal{M}_{str}(h^t, \mathbf{A})$
            \STATE $\hat{\mathbf{X}}_{t+1:t+H} = f_{out}(h^{str})$
            \STATE Compute $\mathcal{L}_{\text{Stage-I}}$ via Eq. (\ref{eq:loss_1})
            \STATE Update $\mathcal{M}_t, \mathcal{M}_{str}, f_{out}$ by minimizing $\mathcal{L}_{\text{Stage-I}}$
        \ENDWHILE
        \STATE Extract and freeze the optimized DDG matrix $G^*$

        \STATE \hrulefill
        \STATE \textbf{Stage II: Soft Alignment of Spatiotemporal Attention}
        \WHILE{Stage II not converged}
            \STATE $h^t = \mathcal{M}_t(\mathbf{X}_{t-T+1:t})$
            \STATE $h^{att} = \mathcal{M}_s(h^t, G^*)$ \COMMENT{Guided by the stable structural prior}
            \STATE $\hat{\mathbf{X}}_{t+1:t+H} = f_{out}(h^{att})$
            \STATE Compute $\mathcal{L}_{\text{Stage-II}}$ via Eq. (\ref{eq:loss_2})
            \STATE Update $\mathcal{M}_t, \mathcal{M}_s, f_{out}$ by minimizing $\mathcal{L}_{\text{Stage-II}}$
        \ENDWHILE
    \end{algorithmic}
\end{algorithm}

The training complexity of our model is reduced through its decoupled design. In Stage I, generating the DDG involves continuous DAG optimization, which carries an $O(N^3)$ complexity. However, this functions strictly as an offline pre-training process. In Stage II, the standard spatial attention complexity of $O(T \cdot N^2 \cdot D)$ is significantly reduced. By masking the attention map with the learned sparse dependency graph, the complexity drops to $O(T \cdot |E| \cdot D)$, where $|E|$ is the number of retained edges ($|E| \ll N^2$). 

During the validation and online testing stages, the structural discovery module is bypassed entirely. The future traffic states are predicted by feeding the dynamic inputs through the spatiotemporal attention module, which is guided purely by the frozen, pre-computed sparse DDG mask. Therefore, the online inference complexity remains strictly bounded at $O(T \cdot |E| \cdot D)$, bypassing the heavy structural generation overhead entirely.

\section{Experiment and Results}
\label{sec:experiment}
To validate our proposed model, we implement and train the model on several real-world traffic flow datasets, compare prediction results with baseline models, and conduct ablation studies on the structural discovery module and the two-stage learning framework. In addition, we compare the training and inference efficiency of these models theoretically and empirically. 

\subsection{Datasets and Experiment Setup}
We adopt four open-source traffic datasets for evaluations, which are provided by the Performance Measurement System (PeMS) of the California Department of Transportation: PeMS03, PeMS07, PeMS08, and PeMS-Bay. Detailed descriptions of the datasets are provided in Table~\ref{tab:datasets}.

\begin{table}[htbp]
  \centering
  \caption{Descriptive Analysis of Datasets}
    \begin{tabular}{ccccc}
    \toprule
    Dataset & PeMS03 & PeMS07 & PeMS08 & PeMS-Bay \\
    \midrule
    Timesteps & 26,208 & 28,224 & 17,856 & 52,116 \\
    Number of nodes & 358 & 883 & 170 & 325 \\
    Number of edges & 549 & 865 & 276 & 2,369 \\
    Predicted state & Volume & Volume & Volume & Mean Speed \\
    \bottomrule
    \end{tabular}
  \label{tab:datasets}
\end{table}

The length of the input sequence is 15 minutes, and the prediction horizon is set to 15, 30, and 60 minutes. The training, validation, and test sets were split in a ratio of 7:1:2. The Adam optimizer is adopted, and the learning rate is $4 \times 10^{-4}$ with a dropout rate of 0.1. The model was trained on an NVIDIA Tesla A800 GPU for 100 epochs, with the first 50 epochs for Stage I and 50 epochs for Stage II to balance training efficiency and the sufficiency of dependency structure learning.

Through hyperparameter tuning, we determined the optimal settings:  the hidden layer dimension in our SGSAN model is set to 16 for every module, and each temporal attention module has 4 heads. The values for $\gamma_{D}$, $\gamma_{S}$ and $\gamma_{I}$ in Equations (\ref{eq:loss_1}) and (\ref{eq:loss_2}) are 0.1, $1 \times 10^{-6}$, and $1 \times 10^{-3}$, respectively. 

To validate the predictive performance of the interpretable model proposed in this paper, we conduct comparative experiments with other baseline models on the same dataset. To clarify, SGSAN (Ours) denotes the proposed model trained with the two-stage framework. 

\begin{itemize}[leftmargin=*]
\item \textbf{Temporal Graph Convolutional Network (T-GCN)} \cite{Zhao}: Combines GCNs for spatial topology with GRUs for temporal dynamics.

\item \textbf{Spatio-Temporal Graph Convolutional Network (STGCN)} \cite{YuYin}: Integrates 1D causal convolutions and graph convolutions without relying on RNNs.

\item \textbf{Graph Multi-Attention Network (GMAN)} \cite{zheng2020}: Employs multiple spatial and temporal attention mechanisms with a transform attention layer.

\item \textbf{Attention-based Spatiotemporal Graph Convolutional Network (ASTGCN)} \cite{Guo}: Incorporates spatial and temporal attention mechanisms within a GCN framework.

\item \textbf{Spatiotemporal Transformer Network (STTN)} \cite{Chen}: Dynamically models dependencies utilizing self-attention mechanisms across spatial and temporal domains.

\item \textbf{Dynamic Spatial-Temporal Trend Transformer (DST2former)} \cite{Chen2025}: Captures spatio-temporal correlations through adaptive embedding and fuses dynamic trends with static graph attributes via Cross Spatial-Temporal Attention.

\item \textbf{Dual Cross-Scale Transformer (DCST)} \cite{Zhou2024}: Utilizes a dual-path architecture to capture dependencies across micro and macro temporal scales.

\item \textbf{Causal Conditional Hidden Markov Model (CCHMM)} \cite{Zhao2023}: Employs mutually supervised prior and posterior networks to disentangle causal representations of physical concepts in multimodal traffic.

\item \textbf{Principal Spatio-Temporal Causal Graph Convolutional Network (PSTCGCN)} \cite{Yang2024}: Leverages causal discovery algorithms to generate a causal graph prior for spatio-temporal forecasting.
\end{itemize}

\subsection{Comparative Analysis of Prediction Accuracy}
\begin{table*}[htbp]
  \centering
  \caption{Performance Comparison: Prediction Accuracy}
    \begin{tabular}{cc|ccc|ccc|ccc|ccc}
    \toprule
    \multirow{2}[2]{*}{Dataset} & \multirow{2}[2]{*}{Model} & \multicolumn{3}{c|}{Horizon=15min} & \multicolumn{3}{c|}{Horizon=30min} & \multicolumn{3}{c|}{Horizon=60min} & \multicolumn{3}{c}{Average} \\
          &       & MAE   & MAPE  & RMSE  & MAE   & MAPE  & RMSE  & MAE   & MAPE  & RMSE  & MAE   & MAPE  & RMSE \\
    \midrule
    \multirow{10}[2]{*}{PeMS03} & T-GCN & 20.18  & 19.13  & 27.82  & 29.02  & 26.65  & 43.36  & 36.42  & 33.86  & 53.03  & 28.54  & 26.55  & 41.40  \\
          & STGCN & 17.05  & 14.41  & 26.76  & 26.96  & 25.99  & 39.58  & 31.72  & 29.73  & 46.78  & 25.24  & 23.38  & 37.71  \\
          & GMAN  & 14.87  & 13.37  & 23.15  & 23.65  & 20.35  & 36.70  & 32.00  & 28.79  & 47.63  & 23.51  & 20.83  & 35.83  \\
          & ASTGCN & 14.40  & 12.26  & 25.17  & 23.98  & 20.68  & 37.71  & 24.07  & 20.25  & 37.65  & 20.81  & 17.73  & 33.51  \\
          & STTN  & 16.43  & 17.38  & 24.20  & 22.40  & 19.12  & 34.14  & 27.38  & 31.77  & 42.68  & 22.07  & 22.75  & 33.67  \\
          & DST2Former & 16.53  & 16.25  & 25.65  & 20.41  & 16.54  & 32.13  & 24.72  & 21.96  & 37.82  & 20.55  & 18.25  & 31.87  \\
          & DCST  & 16.24  & 13.89  & 26.17  & 19.13  & 17.02  & 34.19  & 31.26  & 28.68  & 46.41  & 22.21  & 19.86  & 35.59  \\
          & CCHMM & 23.16  & 19.85  & 35.37  & 33.96  & 31.04  & 50.03  & 28.19  & 27.87  & 41.12  & 28.44  & 26.26  & 42.17  \\
          & PSTCGCN & 16.40  & 13.50  & 24.59  & 19.46  & 15.96  & \textbf{29.69} & \textbf{21.53} & \textbf{17.35} & \textbf{33.25} & 19.13  & 15.60  & 29.18  \\
          & SGSAN (Ours) & \textbf{13.74} & \textbf{11.06} & \textbf{22.56} & \textbf{18.62} & \textbf{15.13} & 30.71  & 22.98  & 18.79  & 36.74  & \textbf{18.45} & \textbf{14.99} & \textbf{30.00} \\
    \midrule
    \multirow{10}[2]{*}{PeMS07} & T-GCN & 33.66  & 15.50  & 47.55  & 38.72  & 18.41  & 54.38  & 53.37  & 28.05  & 73.81  & 41.91  & 20.65  & 58.58  \\
          & STGCN & 29.91  & 12.73  & 44.41  & 36.78  & 16.11  & 53.62  & 47.04  & 23.15  & 65.69  & 37.91  & 17.33  & 54.57  \\
          & GMAN  & 26.02  & 11.47  & 38.51  & 34.37  & 16.64  & 49.35  & 47.11  & 24.90  & 65.07  & 35.83  & 17.67  & 50.97  \\
          & ASTGCN & 23.29  & 10.31  & 35.64  & 31.42  & 13.26  & 46.78  & 37.41  & 17.07  & 53.37  & 30.71  & 13.55  & 45.26  \\
          & STTN  & 23.29  & 11.02  & 35.47  & 36.30  & 16.77  & 49.95  & 43.24  & 19.30  & 57.50  & 34.28  & 15.70  & 47.64  \\
          & DST2Former & 24.92  & 10.94  & 36.50  & 29.34  & 12.37  & 42.15  & 38.58  & 18.82  & 52.99  & 30.94  & 14.05  & 43.88  \\
          & DCST  & 24.34  & 9.91  & 36.25  & 27.99  & 11.51  & 41.12  & 35.78  & 14.62  & 51.14  & 29.37  & 12.01  & 42.83  \\
          & CCHMM & 34.76  & 15.46  & 49.87  & 41.29  & 20.46  & 57.58  & 48.90  & 22.49  & 68.31  & 41.65  & 19.47  & 58.59  \\
          & PSTCGCN & 26.30  & 15.99  & 37.39  & 31.28  & 18.46  & 44.27  & 40.14  & 24.50  & 56.08  & 32.57  & 19.65  & 45.91  \\
          & SGSAN (Ours) & \textbf{22.84} & \textbf{9.31} & \textbf{35.11} & \textbf{27.47} & \textbf{11.05} & \textbf{40.95} & \textbf{33.89} & \textbf{14.75} & \textbf{50.02} & \textbf{28.07} & \textbf{11.70} & \textbf{42.02} \\
    \midrule
    \multirow{10}[2]{*}{PeMS08} & T-GCN & 21.19  & 11.76  & 30.74  & 28.94  & 19.17  & 41.52  & 44.07  & 30.77  & 61.81  & 31.40  & 20.56  & 44.69  \\
          & STGCN & 21.35  & 10.01  & 30.45  & 25.75  & 15.25  & 37.60  & 34.06  & 21.71  & 47.53  & 27.05  & 15.66  & 38.53  \\
          & GMAN  & 15.32  & 9.47  & 26.55  & 25.76  & 16.45  & 37.16  & 34.52  & 22.29  & 48.46  & 25.20  & 16.07  & 37.39  \\
          & ASTGCN & 15.31  & 9.13  & 23.83  & 22.48  & 13.08  & 33.74  & 29.00  & 17.19  & 42.73  & 22.26  & 13.13  & 33.43  \\
          & STTN  & 16.41  & 10.82  & 23.17  & 24.11  & 19.74  & 33.90  & 28.11  & 15.53  & 40.51  & 22.88  & 15.36  & 32.53  \\
          & DST2Former & 17.91  & 9.31  & 25.89  & 21.02  & 12.56  & 31.50  & 26.05  & 16.04  & 37.19  & 21.66  & 12.64  & 31.52  \\
          & DCST  & 16.97  & 9.78  & 25.65  & 21.15  & 12.77  & 32.67  & 25.95  & 15.75  & 40.26  & 21.36  & 12.77  & 32.86  \\
          & CCHMM & 26.80  & 16.51  & 39.91  & 30.69  & 20.33  & 44.49  & 35.67  & 23.52  & 49.72  & 31.05  & 20.12  & 44.71  \\
          & PSTCGCN & 18.43  & 13.67  & 26.70  & 21.89  & 15.83  & 31.72  & 27.65  & 19.91  & 39.66  & 27.65  & 19.91  & 39.66  \\
          & SGSAN (Ours) & \textbf{14.71} & \textbf{8.59} & \textbf{22.04} & \textbf{20.58} & \textbf{11.74} & \textbf{31.25} & \textbf{24.91} & \textbf{14.60} & \textbf{37.18} & \textbf{20.06} & \textbf{11.64} & \textbf{30.15} \\
    \midrule
    \multirow{10}[2]{*}{PeMS-Bay} & T-GCN & 1.82  & 3.74  & 3.40  & 2.42  & 5.28  & 4.71  & 3.00  & 6.87  & 5.74  & 2.41  & 5.30  & 4.62  \\
          & STGCN & 1.78  & 3.49  & 3.08  & 2.60  & 5.66  & 4.91  & 3.33  & 7.53  & 6.11  & 2.57  & 5.56  & 4.70  \\
          & GMAN  & 1.63  & 3.41  & 3.22  & 2.20  & 4.89  & 4.56  & 2.95  & 6.95  & 6.02  & 2.26  & 5.08  & 4.60  \\
          & ASTGCN & 1.54  & 3.19  & 3.18  & 2.10  & 4.61  & 4.42  & 2.77  & 6.10  & 5.81  & 2.14  & 4.63  & 4.47  \\
          & STTN  & 1.64  & 3.39  & 3.18  & 2.26  & 4.64  & 4.20  & 2.82  & 6.17  & 5.11  & 2.24  & 4.73  & 4.16  \\
          & DST2Former & 1.54  & 3.30  & 2.97  & 2.17  & 4.34  & 3.94  & 2.61  & 5.47  & 5.18  & 2.11  & 4.37  & 4.03  \\
          & DCST  & 1.50  & 3.19  & 3.01  & 1.99  & \textbf{4.27} & 3.99  & 2.39  & 5.23  & 4.76  & 1.96  & 4.23  & 3.92  \\
          & CCHMM & 1.66  & 3.43  & 3.23  & 2.26  & 5.05  & 4.53  & 3.13  & 7.31  & 6.02  & 2.35  & 5.26  & 4.59  \\
          & PSTCGCN & 1.93  & 4.80  & 4.26  & 2.45  & 5.96  & 5.15  & 2.95  & 7.38  & 6.17  & 2.45  & 6.05  & 5.19  \\
          & SGSAN (Ours) & \textbf{1.48} & \textbf{3.01} & \textbf{2.70} & \textbf{1.99} & 4.30  & \textbf{3.87} & \textbf{2.38} & \textbf{5.17} & \textbf{4.60} & \textbf{1.95} & \textbf{4.16} & \textbf{3.73} \\
    \bottomrule
    \end{tabular}
  \label{tab:accuracy}
\end{table*}

As shown in Table~\ref{tab:accuracy}, the proposed SGSAN (Ours) method achieves state-of-the-art predictive accuracy on all four datasets. Compared to models with Transformer structures, such as STTN, DST2Former, and DCST, SGSAN (Ours) outperforms them, especially when the prediction horizon is not so long. An interesting phenomenon is that these models with more complicated structures fail to demonstrate their high capacity when the prediction horizon is 15 minutes. Nevertheless, when the prediction horizon increases, the advantages of capturing long sequence dependency are fully realized. For the proposed model SGSAN (Ours), the structural discovery module corrects the potential spurious correlations introduced by unconstrained attention and finds sparser and more global feature significance, further improving the prediction accuracy. This highlights that the Structure-Guided spatiotemporal attention mechanism demonstrates a strong capacity for prediction by identifying the dominant propagation paths and interpreting the built-in mechanism of traffic flow dynamics.

\subsection{Ablation Studies}
\label{subsec:ablation}
The ablation studies examine the effectiveness of the DDG and the two-stage training framework. Table~\ref{tab:ablation} shows the prediction accuracy in the ablation studies, where baseline models used for comparison are
\begin{itemize}[leftmargin=*]
    % \item \textbf{SGSAN (Ours)} -- The proposed two-stage training framework with a structural discovery module. Furthermore, we compare InfoNCE against Mean Squared Error (MSE), Cosine Similarity, and KL divergence to fuse the static DDG with dynamic attention under the identical framework.
    
    \item \textbf{STA-GNN} (Spatiotemporal Attention-based Graph Neural Network) -- The structural discovery module has been entirely removed, and the prediction results are only derived from the attention module.
    
    \item \textbf{SGSAN (DAG-Relaxed):} The model architecture and two-stage training framework remain identical to the proposed SGSAN. We introduce a tolerance margin $\epsilon = 0.1$ into the dependency structural loss $\mathcal{L}_{D}$ to relax the strict acyclicity constraint. This can address the concern that strict DAG assumption may over-constrain the structural learning process.

    \item \textbf{SGSAN (Random-DDG):} The model architecture and two-stage training framework remain identical to the proposed SGSAN. However, rather than employing the DDG $\mathcal{G}^{*}$ learned through the structural discovery module in Stage I, we substitute it with a randomly generated sparse DAG $\mathcal{G}^{\text{rand}}$ of equivalent edge density, whose non-zero entries are drawn from a uniform distribution and subsequently normalized.
    
    \item \textbf{SGSAN (Joint)} -- A single-stage joint training framework. The architecture is the same as the proposed model but trained end-to-end simultaneously with all losses active.

\end{itemize}

The comprehensive ablation results presented in Table ~\ref{tab:ablation} demonstrate the advantages of the proposed SGSAN and validate the contributions of its individual components and the two-stage training framework.

\begin{table*}
  \centering
  \caption{Prediction Results in the Ablation Studies}
        \begin{tabular}{cc|ccc|ccc|ccc}
    \toprule
    \multirow{2}[2]{*}{Dataset} & \multirow{2}[2]{*}{Model} & \multicolumn{3}{c|}{Horizon=15min} & \multicolumn{3}{c|}{Horizon=30min} & \multicolumn{3}{c}{Horizon=60min} \\
          &       & MAE   & MAPE  & RMSE  & MAE   & MAPE  & RMSE  & MAE   & MAPE  & RMSE \\
    \midrule
    \multirow{8}[2]{*}{PeMS03} & STA-GNN & 17.70  & 14.42  & 28.88  & 21.30  & 17.02  & 34.54  & 28.36  & 23.07  & 44.46  \\
          & SGSAN (DAG-Relaxed) & 16.30  & 18.36  & 30.40  & 20.74  & 22.95  & 33.52  & 24.15  & 22.71  & 37.60  \\
          & SGSAN (Random-DDG) & 16.45  & 17.76  & 26.30  & 23.28  & 26.86  & 33.10  & 25.47  & 26.42  & 39.07  \\
          & SGSAN (Joint) & 14.05  & 11.20  & 22.95  & 25.59  & 23.59  & 38.60  & 31.72  & 32.36  & 44.87  \\
          & SGSAN (Ours) - KL & 15.80  & 13.35  & 24.85  & 20.23  & 16.12  & 32.02  & 24.26  & 20.90  & 37.08  \\
          & SGSAN (Ours) - Cosine & 15.57  & 12.03  & 23.57  & 19.91  & 16.82  & 31.98  & 24.22  & 19.78  & 37.02  \\
          & SGSAN (Ours) - MSE & 16.75  & 12.88  & 22.67  & 22.09  & 17.92  & 31.79  & 24.35  & 26.94  & \textbf{36.23} \\
          & SGSAN (Ours) - InfoNCE & \textbf{13.74} & \textbf{11.06} & \textbf{22.56} & \textbf{18.62} & \textbf{15.13} & \textbf{30.71} & \textbf{22.98} & \textbf{18.79} & 36.74  \\
    \midrule
    \multirow{8}[2]{*}{PeMS07} & STA-GNN & 25.52  & 11.65  & 37.73  & 32.65  & 14.51  & 47.61  & 42.16  & 19.32  & 60.06  \\
          & SGSAN (DAG-Relaxed) & 28.87  & 14.38  & 39.23  & 30.86  & 18.51  & 42.89  & 36.16  & 16.30  & 53.73  \\
          & SGSAN (Random-DDG) & 27.46  & 15.21  & 38.34  & 30.81  & 17.71  & 42.29  & 34.33  & 18.74  & 53.52  \\
          & SGSAN (Joint) & 25.06  & 11.35  & 37.20  & 58.93  & 29.93  & 77.35  & 70.26  & 36.53  & 89.04  \\
          & SGSAN (Ours) - KL & 28.36  & 11.70  & 38.39  & 30.79  & 14.54  & 44.04  & 34.99  & 17.97  & 52.25  \\
          & SGSAN (Ours) - Cosine & 24.56  & 13.61  & \textbf{34.67} & 30.89  & 15.22  & 42.31  & 35.76  & 17.93  & 52.18  \\
          & SGSAN (Ours) - MSE & 28.40  & 11.50  & 38.62  & 31.46  & 14.29  & 42.61  & \textbf{33.57} & 16.49  & 51.39  \\
          & SGSAN (Ours) - InfoNCE & \textbf{22.84} & \textbf{9.31} & 35.11  & \textbf{27.47} & \textbf{11.05} & \textbf{40.95} & 33.89  & \textbf{14.75} & \textbf{50.02} \\
    \midrule
    \multirow{8}[2]{*}{PeMS08} & STA-GNN & 18.55  & 10.64  & 28.15  & 22.26  & 12.56  & 33.78  & 30.22  & 17.49  & 43.82  \\
          & SGSAN (DAG-Relaxed) & 16.32  & 9.54  & 25.89  & 23.31  & 16.66  & 32.25  & 26.70  & 18.89  & 39.59  \\
          & SGSAN (Random-DDG) & 19.92  & 10.42  & 27.31  & 24.28  & 17.84  & 32.96  & 25.92  & 18.46  & 41.25  \\
          & SGSAN (Joint) & 15.75  & 9.78  & 23.13  & 24.92  & 14.62  & 37.53  & 26.51  & 16.19  & 39.34  \\
          & SGSAN (Ours) - KL & 15.92  & 9.50  & 23.79  & 22.37  & 12.34  & 32.06  & 25.96  & 18.05  & 38.89  \\
          & SGSAN (Ours) - Cosine & 15.33  & 9.19  & 22.93  & 22.37  & 13.14  & 33.14  & 26.23  & 17.12  & 38.11  \\
          & SGSAN (Ours) - MSE & 15.44  & 9.28  & 23.12  & 22.41  & 12.49  & 32.07  & 25.78  & 18.66  & 38.49  \\
          & SGSAN (Ours) - InfoNCE & \textbf{14.71} & \textbf{8.59} & \textbf{22.04} & \textbf{20.58} & \textbf{11.74} & \textbf{31.25} & \textbf{24.91} & \textbf{14.60} & \textbf{37.18} \\
    \midrule
    \multirow{8}[2]{*}{PeMS-Bay} & STA-GNN & 1.51  & 3.08  & 3.11  & 2.07  & 4.59  & 4.40  & 2.83  & 6.74  & 5.89  \\
          & SGSAN (DAG-Relaxed) & 1.62  & 3.59  & 3.07  & 2.03  & 4.46  & 3.92  & 2.42  & 5.34  & 4.66  \\
          & SGSAN (Random-DDG) & 1.61  & 3.47  & 3.01  & 2.02  & 4.47  & 3.93  & 2.39  & 5.34  & 4.63  \\
          & SGSAN (Joint) & 2.49  & 3.75  & 3.84  & 3.30  & 6.54  & 5.65  & 3.86  & 7.02  & 5.30  \\
          & SGSAN (Ours) - KL & 1.62  & 3.56  & 3.05  & 2.00  & 4.46  & 3.91  & \textbf{2.37} & 5.34  & 4.66  \\
          & SGSAN (Ours) - Cosine & 1.62  & 3.55  & 3.04  & 2.00  & 4.43  & 3.90  & 2.37  & 5.39  & 4.68  \\
          & SGSAN (Ours) - MSE & 1.55  & 3.28  & 2.87  & 1.99  & 4.36  & 3.89  & 2.37  & 5.28  & 4.64  \\
          & SGSAN (Ours) - InfoNCE & \textbf{1.48} & \textbf{3.01} & \textbf{2.70} & \textbf{1.99} & \textbf{4.30} & \textbf{3.87} & 2.38  & \textbf{5.17} & \textbf{4.60} \\
    \bottomrule
    \end{tabular}%
  \label{tab:ablation}%
\end{table*}

Compared to the proposed two-stage framework, SGSAN (Joint) struggles to converge on complex networks. For example, its MAE on PeMS07 at the 60-min horizon increased sharply to 70.26. This indicates that simultaneously learning macroscopic priors and local correlations leads to severe optimization conflicts, a bottleneck our decoupled framework effectively resolves.

To rule out that improvements stem solely from sparsity penalties, we evaluate SGSAN (Random-DDG). Applying a random sparse prior without structural guidance sometimes leads to worse performance than the unconstrained STA-GNN (e.g., PeMS07 at the 15-min horizon). Furthermore, the degradation of SGSAN (DAG-Relaxed) shows that relaxing strict acyclic constraints introduces spurious cyclic noise. The marked advantage of SGSAN (Ours) over these baselines validates that our framework successfully captures genuine macroscopic dependencies.

Among coupling mechanisms, InfoNCE outperforms MSE, cosine, and KL divergence in many cases. Hard constraints force the dynamic attention matrix to rigidly replicate the static DDG, restricting its capability to adapt to real-time, non-topological local fluctuations. In contrast, InfoNCE anchors the attention mechanism without forcing absolute equality, granting the numerical freedom to capture dynamic local variances. Admittedly, strict numerical regularization provides tight bounds under specific conditions, occasionally giving alternative metrics a slight edge. However, InfoNCE still holds the advantage across MAE, MAPE, and RMSE in general.

\subsection{Computational Efficiency}
\label{subsec:efficiency}

In terms of computational efficiency, the proposed two-stage framework excels in both the training and deployment phases. We evaluate its performance from two perspectives: training speed and online inference latency.

\subsubsection{Training Efficiency}
As shown in Table~\ref{tab:training_time}, this two-stage decoupled approach SGSAN (Ours) reduces the average training time per epoch by approximately 31.7\% across all datasets compared to the joint-learning baseline SGSAN (Joint), effectively mitigating the computational bottleneck of simultaneous graph generation and prediction.

\begin{table}[htbp]
  \centering
  \caption{Average Training Time Per Epoch}
  \begin{tabular}{lcccc}
    \toprule
    Model & PeMS03 & PeMS07 & PeMS08 & PeMS-Bay \\
    \midrule
    SGSAN (Joint) & 11.35s & 52.21s & 4.63s  & 21.20s \\
    SGSAN (Ours) & 8.35s  & 31.73s & 3.25s  & 14.57s \\
    \midrule
    Reduction & -26.4\% & -39.2\% & -29.8\% & -31.3\% \\
    \bottomrule
  \end{tabular}
  \label{tab:training_time}
\end{table}

\subsubsection{Inference Efficiency}
For real-world urban computing deployments, inference efficiency is a crucial performance metric. Because our global dependency graph is statically learned offline, it introduces zero structural generation overhead during the online inference phase. The online data flow is comparable to standard GNNs, requiring only the application of the pre-learned sparse graph mask.

Table~\ref{tab:inference_time} compares the average inference time per sample on the PeMS03 dataset. SGSAN (Ours) achieves a low latency of 0.096 ms/sample. Notably, it is vastly more efficient than the recent causal baseline PSTCGCN, which relies on heavier online dependency structure generation. While ASTGCN and CCHMM are marginally faster, our framework offers a significantly superior trade-off by achieving much higher predictive accuracy, making it highly suitable for real-time, safety-critical traffic management systems. Additionally, the proposed SGSAN (Ours) is considerably faster than the baseline SGSAN (Joint).

\begin{table}[htbp]
  \centering
  \caption{Inference Efficiency Comparison}
  \begin{tabular}{lc}
    \toprule
    Model & Avg Time per Sample (ms) \\
    \midrule
    T-GCN & 0.044 $\pm$ 0.003 \\
    ASTGCN & 0.078 $\pm$ 0.001 \\
    CCHMM & 0.079 $\pm$ 0.004 \\
    \textbf{SGSAN (Ours)} & 0.096 $\pm$ 0.003 \\
    DCST & 0.125 $\pm$ 0.001 \\
    STCGCN & 0.136 $\pm$ 0.002 \\
    DST2Former & 0.155 $\pm$ 0.002 \\
    PSTCGCN & 0.434 $\pm$ 0.008 \\
    SGSAN (Joint) & 0.511 $\pm$ 0.005 \\
    STTN & 0.513 $\pm$ 0.001 \\

    \bottomrule
  \end{tabular}
  \label{tab:inference_time}
\end{table}

\subsection{Interpretability Analysis}
\label{subsec:interpretability}

\begin{figure*}[h]
    \centering
    \includegraphics[width=\linewidth]{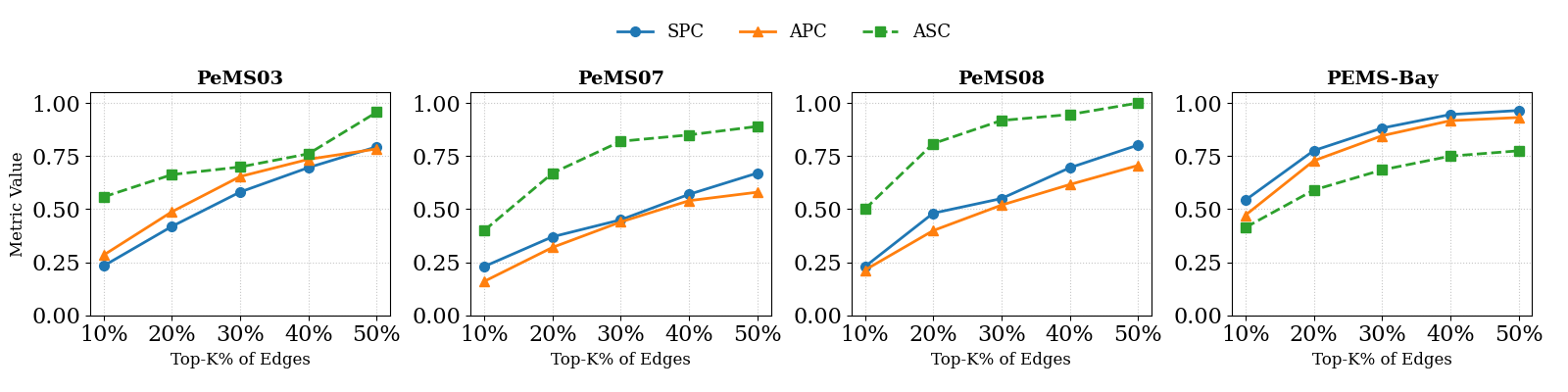}
    \caption{Sensitivity Analysis of Interpretability Metrics (ASC, SPC, and APC) across varying sparsity levels (Top-$K\%$)}
    \label{fig:topk}
\end{figure*}

\begin{figure*}[h]
    \centering
    \subfloat[Example 1]{\includegraphics[width=0.48\linewidth]{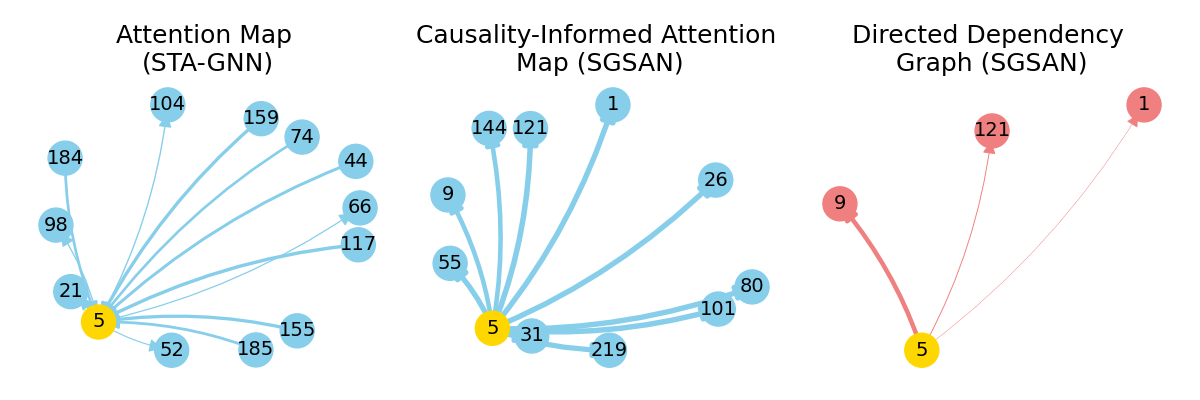}}
    \hfill
    \subfloat[Example 2]{\includegraphics[width=0.48\linewidth]{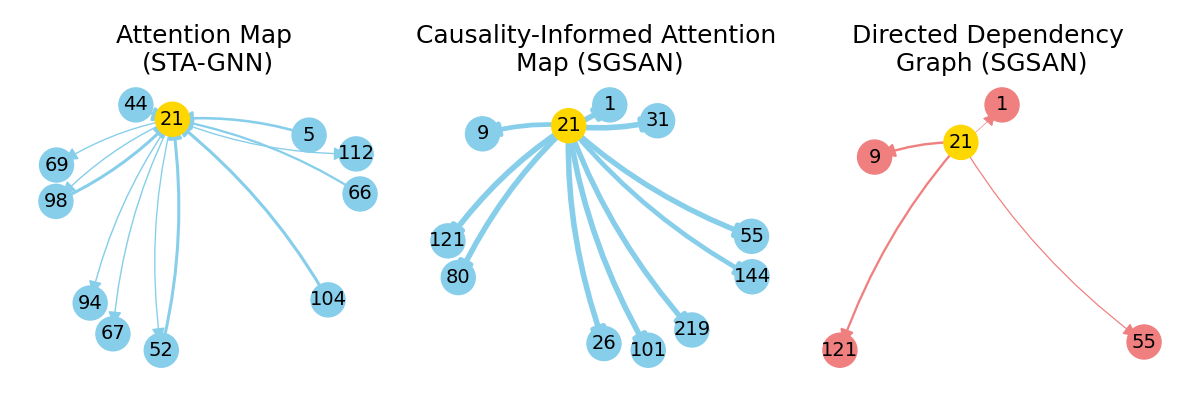}}
    \\
    \subfloat[Example 3]{\includegraphics[width=0.48\linewidth]{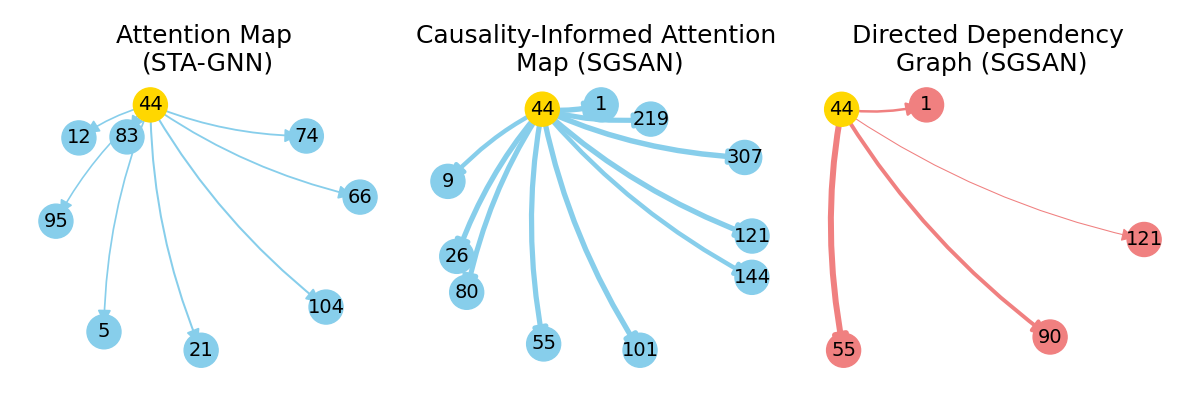}}
    \hfill
    \subfloat[Example 4]{\includegraphics[width=0.48\linewidth]{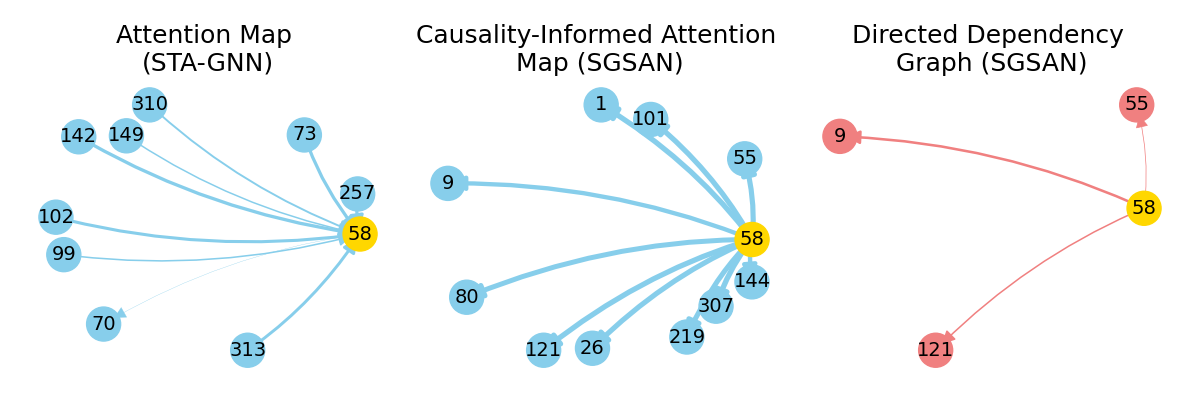}}
    \\
    \subfloat[Example 5]{\includegraphics[width=0.48\linewidth]{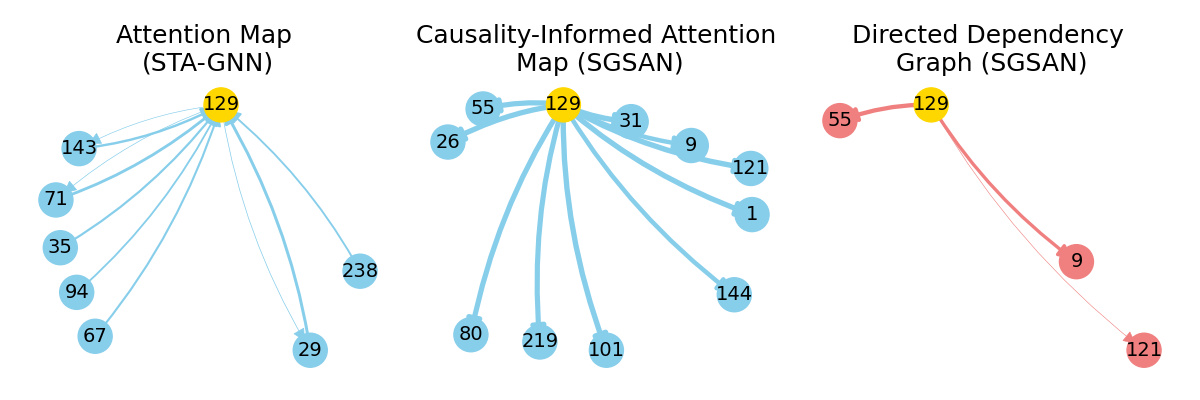}}
    \hfill
    \subfloat[Example 6]{\includegraphics[width=0.48\linewidth]{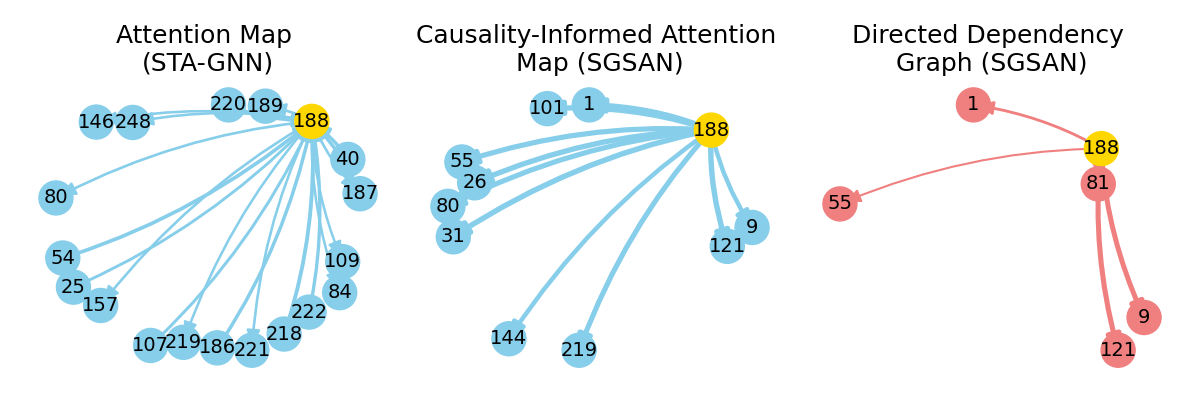}}
    
    \caption{Example Nodes in the Subgraphs illustrating consistency between the DDG and the Attention Map.}
    \label{fig:subgraph_detailed}
\end{figure*}

To comprehensively demonstrate that SGSAN provides trustworthy and interpretable prediction, we conduct both qualitative and quantitative analyses on its interpretability. We evaluate internal consistency to show that the dynamic attention is properly anchored to the learned structural backbone, and we evaluate external physical alignment to verify that the learned components naturally reflect the real-world road network topology. Additionally, we examine dynamic flexibility to ensure the model adaptively balances global structural reliance with local stochasticity.

\subsubsection{Consistency Metrics for Structure-Guided Interpretability}
\label{subsec:inter_metric}
To quantitatively validate the structure-guided mechanisms of SGSAN, we evaluate the alignment between the structural dependency prior and the dynamic spatiotemporal attention. A foundational assumption of our framework is that a truly interpretable and trustworthy model should implicitly capture the physical macroscopic backbone of the road network, rather than relying on spurious statistical noise.

\textbf{(i) Attention-Structural Consistency (ASC)}.
This metric measures the internal alignment between the static DDG and the dynamic spatiotemporal attention map. The rationale is that a trustworthy model should anchor its attention weights to the underlying structural backbone identified in Stage I. A high ASC indicates the effectiveness of the InfoNCE contrastive learning loss. 

\begin{equation}
    \label{eq:asc}
    ASC = \frac{1}{B}\sum_{b=1}^{B} \frac{1}{K}|\{\mathbf{TopK}(G^a(b))\} \cap \{\mathbf{TopK}(G^*)\}|
\end{equation}

where $\mathbf{TopK}(\cdot)$ denotes the operation of selecting the top $K\%$ of node pairs with the highest weight values. $G^a(b)$ denotes the attention map associated with batch $b$, and $B$ denotes the total number of batches. While the dependency graph ($G^*$) remains static, the attention map dynamically adapts to real-time inputs in each batch. A higher $ASC$ value indicates a stronger structural prior guiding the attention mechanism.

\textbf{(ii) Structural-Physical Consistency (SPC)} and \textbf{Attention-Physical Consistency (APC)}.
While ASC measures internal alignment, there is a risk that the metric is self-referential, as Stage II explicitly integrates the dependency graph and attention maps. To provide an external evaluation, we introduce SPC and APC. SPC measures the overlap between the learned DDG and the actual physical road network adjacency matrix, while APC measures the overlap between the dynamic attention map and the physical network.

\subsubsection{Quantitative Analysis of Structural Consistency}
This section presents the quantitative results of interpretability using the metrics introduced in Section \ref{subsec:inter_metric}.

Figure~\ref{fig:topk} presents a sensitivity analysis of these metrics across varying sparsity levels (from Top-10\% to Top-50\% retained capacity) for all four datasets. The results yield three crucial insights regarding the model's interpretability.

First, the high SPC scores across datasets demonstrate that our learned dependency graph organically aligns with the physical topology, successfully capturing the true macroscopic structure of the traffic network without any geographic supervision. 

Moreover, the APC and SPC curves exhibit closely matched values and parallel trajectories across all sparsity levels (e.g., at the Top-30\% threshold on PeMS-Bay, SPC and APC reach 0.88 and 0.85, respectively). This proves that the dynamic attention mechanism faithfully inherits the physical alignment of the structural prior.

Finally, the ASC metric proves highly robust across varying network densities. The InfoNCE coupling effectively anchors the attention to the core structural skeleton while retaining the numerical freedom required for the attention mechanism to capture real-time, non-topological dependencies. 

\subsubsection{Visualization of Graph Alignment}
% \begin{figure*}
%     \centering
%     \subfloat[Attention Subgraph]{
%     \includegraphics[width=0.75\linewidth]{attention_map_pemsbay.png}}
%     \\
%     \subfloat[DDG Subgraph]{
%     \includegraphics[width=0.75\linewidth]{causal_graph_pemsbay.png}
%}
%     \caption{Key subgraphs of the Attention Map and DDG (PeMS-Bay dataset). Only the subgraphs composed of the first 60 nodes are displayed.}
%     \label{fig:subgraph_1}
% \end{figure*}

This section offers a qualitative analysis of the model's interpretability, highlighting the visual alignment between the structural dependencies and the attention mechanism. We utilize the sample-specific attention map $G^a$ along with the static dependency graph $G^*$ to identify and filter significant node pairs, creating "key subgraphs" for visualization. While the dependency graph remains uniform across all inputs, the attention map varies dynamically. For this visualization, we average the attention weights across the test set and juxtapose them with the dependency graph. The edges in the key subgraphs represent the most significant node-to-node dependencies.

% Figure~\ref{fig:subgraph_1} illustrates the key subgraphs of both matrices. Generally, the distribution of attention weights is more dispersed compared to the DDG. The DDG emphasizes only the most dominant, structural relationships, rendering it sparser with fewer highly active edges. This is a direct result of the continuous DAG optimization and the sparsity loss ($\mathcal{L}_S$) applied during Stage I training.

Figure~\ref{fig:subgraph_detailed} demonstrates the relationship dynamics for six typical nodes. A strong consistency between the directed dependencies and the attention-based correlations is evident, with the width of each arrow indicating the weight of the relationship. 

A key observation is that while the dependency graph is sparser and highly focused, the majority of the dominant paths it identifies are faithfully mirrored in the attention map. Because the dependency graph learns from the macroscopic structure of the entire dataset, it provides the model with a global perspective to identify the true propagation paths of traffic flow. 

When comparing this Structure-Guided attention map with the baseline STA-GNN, a critical difference emerges. Without structural constraints, standard attention weights tend to be symmetric and undirected, often producing bidirectional arrows that merely reflect statistical correlation rather than physical influence. In contrast, because our dependency graph is constrained to be directed and acyclic, the Structure-Guided attention map adopts these asymmetric relationships. Bidirectional arrows rarely appear, demonstrating that the InfoNCE contrastive learning successfully transfers the directional logic of the physical network into the attention mechanism. SGSAN aligns its prediction process with the actual physical features governing traffic state propagation, thereby making it trustworthy and robust.

\section{Conclusion}
\label{sec:conclusion}

In this paper, we proposed the Structure-Guided Spatiotemporal Attention Graph Neural Network (SGSAN) to bridge the gap between high-accuracy deep learning and trustworthy traffic flow prediction. Traditional attention mechanisms are fundamentally driven by observational correlations, making them susceptible to local noise and spurious correlations. To address this limitation, our framework explicitly decouples structural discovery from representation learning via a two-stage training strategy. By learning a time-invariant Directed Dependency Graph (DDG) and employing an InfoNCE-based soft-coupling mechanism, SGSAN successfully anchors dynamic spatiotemporal attention to a stable structural prior.

Extensive experiments on four real-world datasets demonstrate that SGSAN achieves state-of-the-art predictive performance. In terms of computational efficiency, SGSAN (Ours) outperforms most baselines in inference efficiency, while the two-stage training framework enhances the training efficiency. Furthermore, through newly introduced structural consistency metrics, we verify that the model provides built-in interpretability. The learned DDG achieves an 88\% alignment with the physical structure of real networks, without relying on any geographic constraints. By shifting the focus from unconstrained statistical correlations to directed structural dependencies, this work provides a trustworthy, highly accurate, and deployable modeling framework for transportation systems.

\FloatBarrier

\appendices
\section*{Acknowledgment}
This research was supported by [2024YFB4303100, National Key R\&D Program of China] and [52402407, the National Natural Science Foundation of China].

\bibliographystyle{IEEEtran}
\bibliography{reference}

@inproceedings{Zheng,
 author = {Zheng, Xun and Aragam, Bryon and Ravikumar, Pradeep K and Xing, Eric P},
 booktitle = {Advances in Neural Information Processing Systems},
 editor = {S. Bengio and H. Wallach and H. Larochelle and K. Grauman and N. Cesa-Bianchi and R. Garnett},
 pages = {},
 publisher = {Curran Associates, Inc.},
 title = {DAGs with NO TEARS: Continuous Optimization for Structure Learning},
 volume = {31},
 year = {2018}
}

@article{Do,
   author = {Do, L. N. N. and Vu, H. L. and Vo, B. Q. and Liu, Z. Y. and Phung, D.},
   title = {An effective spatial-temporal attention based neural network for traffic flow prediction},
   journal = {Transportation Research Part C-Emerging Technologies},
   volume = {108},
   pages = {12-28},
   ISSN = {0968-090x},
      year = {2019},
   type = {Journal Article}
}

@article{Guo,
   author = {Guo, S. N. and Lin, Y. F. and Feng, N. and Song, C. and Wan, H. Y.},
   title = {Attention Based Spatial-Temporal Graph Convolutional Networks for Traffic Flow Forecasting},
   journal = {Thirty-Third Aaai Conference on Artificial Intelligence / Thirty-First Innovative Applications of Artificial Intelligence Conference / Ninth Aaai Symposium on Educational Advances in Artificial Intelligence},
   pages = {922-929},
   ISSN = {2159-5399},
   year = {2019},
   type = {Journal Article}
}

@article{Wang,
   author = {Wang, C. X. and Liang, Y. X. and Tan, G. R.},
   title = {wCityCAN: Causal Attention Network for Citywide Spatio-Temporal Forecasting},
   journal = {Proceedings of the 17th Acm International Conference on Web Search and Data Mining, Wsdm 2024},
   pages = {702-711},
      year = {2024},
   type = {Journal Article}
}

@inproceedings{zheng2020,
  title={Gman: A graph multi-attention network for traffic prediction},
  author={Zheng, Chuanpan and Fan, Xiaoliang and Wang, Cheng and Qi, Jianzhong},
  booktitle={Proceedings of the AAAI conference on artificial intelligence},
  volume={34},
  number={01},
  pages={1234--1241},
  year={2020}
}

@article{ZhangTino,
   author = {Zhang, Y. and Tino, P. and Leonardis, A. and Tang, K.},
   title = {A Survey on Neural Network Interpretability},
   journal = {Ieee Transactions on Emerging Topics in Computational Intelligence},
   volume = {5},
   number = {5},
   pages = {726-742},
   ISSN = {2471-285x},
      year = {2021},
   type = {Journal Article}
}

@article{Zhao,
   author = {Zhao, L. and Song, Y. J. and Zhang, C. and Liu, Y. and Wang, P. and Lin, T. and Deng, M. and Li, H. F.},
   title = {T-GCN: A Temporal Graph Convolutional Network for Traffic Prediction},
   journal = {Ieee Transactions on Intelligent Transportation Systems},
   volume = {21},
   number = {9},
   pages = {3848-3858},
   ISSN = {1524-9050},
      year = {2020},
   type = {Journal Article}
}

@article{Oord,
       author = {{van den Oord}, Aaron and {Li}, Yazhe and {Vinyals}, Oriol},
        title = "{Representation Learning with Contrastive Predictive Coding}",
      journal = {arXiv e-prints},
         year = 2018,
        month = jul,
          eid = {arXiv:1807.03748},
        pages = {arXiv:1807.03748},
          }

@InProceedings{Yu,
  title = 	 {{DAG}-{GNN}: {DAG} Structure Learning with Graph Neural Networks},
  author =       {Yu, Yue and Chen, Jie and Gao, Tian and Yu, Mo},
  booktitle = 	 {Proceedings of the 36th International Conference on Machine Learning},
  pages = 	 {7154--7163},
  year = 	 {2019},
  editor = 	 {Chaudhuri, Kamalika and Salakhutdinov, Ruslan},
  volume = 	 {97},
  series = 	 {Proceedings of Machine Learning Research},
  month = 	 {09--15 Jun},
  publisher =    {PMLR}
}

@inproceedings{YuYin,
   Author = {Yu, Bing and Yin, Haoteng and Zhu, Zhanxing},
   Title = {Spatio-Temporal Graph Convolutional Networks: A Deep Learning Framework for Traffic Forecasting},
   Booktitle = {Proceedings of the 27th International Joint Conference on Artificial Intelligence (IJCAI)},
   Year = {2018},
   Pages = {3634-3640}
}

@article{Chen,
   author = {Chen, C. L. and Liu, Y. B. and Chen, L. and Zhang, C. Q.},
   title = {Bidirectional Spatial-Temporal Adaptive Transformer for Urban Traffic Flow Forecasting},
   journal = {Ieee Transactions on Neural Networks and Learning Systems},
   volume = {34},
   number = {10},
   pages = {6913-6925},
   ISSN = {2162-237x},
   year = {2023},
   type = {Journal Article}
}

@article{Cui,
   author = {Cui, Z. Y. and Henrickson, K. and Ke, R. M. and Wang, Y. H.},
   title = {Traffic Graph Convolutional Recurrent Neural Network: A Deep Learning Framework for Network-Scale Traffic Learning and Forecasting},
   journal = {Ieee Transactions on Intelligent Transportation Systems},
   volume = {21},
   number = {11},
   pages = {4883-4894},
   ISSN = {1524-9050},
      year = {2020},
   type = {Journal Article}
}

@article{Zhai,
  title={A Novel Interpretability Evaluation Framework to Understand Deep Learning Traffic Prediction Models},
  author={Xuehao Zhai and Fangce Guo and Aruna Sivakumar},
  journal={SSRN Electronic Journal},
  year={2022},
}

@article{Medrano,
   author = {de Medrano, R. and Aznarte, J. L.},
   title = {A spatio-temporal attention-based spot-forecasting framework for urban traffic prediction},
   journal = {Applied Soft Computing},
   volume = {96},
   ISSN = {1568-4946},
   year = {2020},
   type = {Journal Article}
}

@article{Ribeiro,
  title={"Why Should I Trust You?": Explaining the Predictions of Any Classifier},
  author={ Ribeiro, Marco Tulio  and  Singh, Sameer  and  Guestrin, Carlos },
  journal={ACM},
  year={2016},
}

@article{Bai,
   author = {Bai, J. D. and Zhu, J. W. and Song, Y. J. and Zhao, L. and Hou, Z. X. and Du, R. H. and Li, H. F.},
   title = {A3T-GCN: Attention Temporal Graph Convolutional Network for Traffic Forecasting},
   journal = {Isprs International Journal of Geo-Information},
   volume = {10},
   number = {7},
   year = {2021},
   type = {Journal Article}
}

@article{Lan,
   author = {Lan, S. Y. and Ma, Y. T. and Huang, W. K. and Wang, W. W. and Yang, H. Y. and Li, P. Y.},
   title = {DSTAGNN: Dynamic Spatial-Temporal Aware Graph Neural Network for Traffic Flow Forecasting},
   journal = {International Conference on Machine Learning, Vol 162},
   ISSN = {2640-3498},
   year = {2022},
   type = {Journal Article}
}

@article{Li,
   author = {Li, G. P. and Knoop, V. L. and van Lint, H.},
   title = {Multistep traffic forecasting by dynamic graph convolution: Interpretations of real-time spatial correlations},
   journal = {Transportation Research Part C-Emerging Technologies},
   volume = {128},
   ISSN = {0968-090x},
   year = {2021},
   type = {Journal Article}
}

@article{Tygesen,
   author = {Tygesen, M. N. and Pereira, F. C. and Rodrigues, F.},
   title = {Unboxing the graph: Towards interpretable graph neural networks for transport prediction through neural relational inference},
   journal = {Transportation Research Part C-Emerging Technologies},
   volume = {146},
   ISSN = {0968-090x},
   year = {2023},
   type = {Journal Article}
}

@article{ZhangZheng,
   author = {Zhang, K. P. and Zheng, L. and Liu, Z. J. and Jia, N.},
   title = {A deep learning based multitask model for network-wide traffic speed prediction},
   journal = {Neurocomputing},
   volume = {396},
   pages = {438-450},
   ISSN = {0925-2312},
      year = {2020},
   type = {Journal Article}
}

@article{Vlahogianni,
   author = {Vlahogianni, E. I. and Karlaftis, M. G. and Golias, J. C.},
   title = {Short-term traffic forecasting: Where we are and where we're going},
   journal = {Transportation Research Part C-Emerging Technologies},
   volume = {43},
   pages = {3-19},
   ISSN = {0968-090x},
      year = {2014},
   type = {Journal Article}
}

@article{Lv,
  author={Lv, Yisheng and Duan, Yanjie and Kang, Wenwen and Li, Zhengxi and Wang, Fei-Yue},
  journal={IEEE Transactions on Intelligent Transportation Systems}, 
  title={Traffic Flow Prediction With Big Data: A Deep Learning Approach}, 
  year={2015},
  volume={16},
  number={2},
  pages={865-873},
}

@article{Janzing,
   author = {Janzing, D.},
   title = {Causal Regularization},
   journal = {Advances in Neural Information Processing Systems 32 (Nips 2019)},
   volume = {32},
   ISSN = {1049-5258},
   year = {2019},
   type = {Journal Article}
}

@article{Tian,
   author = {Tian, Y. X. and Pan, L.},
   title = {Predicting Short-term Traffic Flow by Long Short-Term Memory Recurrent Neural Network},
   journal = {2015 Ieee International Conference on Smart City/Socialcom/Sustaincom (Smartcity)},
   pages = {153-158},
      year = {2015},
   type = {Journal Article}
}

@article{Liang,
   author = {Liang, Y. and Li, S. G. and Yan, C. G. and Li, M. Z. and Jiang, C. J.},
   title = {Explaining the black-box model: A survey of local interpretation methods for deep neural networks},
   journal = {Neurocomputing},
   volume = {419},
   pages = {168-182},
   ISSN = {0925-2312},
      year = {2021},
   type = {Journal Article}
}

@article{Niu,
   author = {Niu, Z. Y. and Zhong, G. Q. and Yu, H.},
   title = {A review on the attention mechanism of deep learning},
   journal = {Neurocomputing},
   volume = {452},
   pages = {48-62},
   ISSN = {0925-2312},
      year = {2021},
   type = {Journal Article}
}

@article{Ali,
   author = {Ali, A. and Zhu, Y. M. and Zakarya, M.},
   title = {Exploiting dynamic spatio-temporal correlations for citywide traffic flow prediction using attention based neural networks},
   journal = {Information Sciences},
   volume = {577},
   pages = {852-870},
   ISSN = {0020-0255},
      year = {2021},
   type = {Journal Article}
}

@article{Arrieta,
   author = {Arrieta, A. B. and Díaz-Rodríguez, N. and Del Ser, J. and Bennetot, A. and Tabik, S. and Barbado, A. and García, S. and Gil-López, S. and Molina, D. and Benjamins, R. and Chatila, R. and Herrera, F.},
   title = {Explainable Artificial Intelligence (XAI): Concepts, taxonomies, opportunities and challenges toward responsible AI},
   journal = {Information Fusion},
   volume = {58},
   pages = {82-115},
   ISSN = {1566-2535},
      year = {2020},
   type = {Journal Article}
}

@book{Pearl,
author = {Pearl, Judea},
title = {Causality: models, reasoning, and inference},
year = {2000},
isbn = {0521773628},
publisher = {Cambridge University Press},
address = {USA}
}

@article{Yin,
  title={Deep learning on traffic prediction: Methods, analysis, and future directions},
  author={Yin, Xueyan and Wu, Genze and Wei, Jinze and Shen, Yanming and Qi, Heng and Yin, Baocai},
  journal={IEEE Transactions on Intelligent Transportation Systems},
  volume={23},
  number={6},
  pages={4927--4943},
  year={2021},
  publisher={IEEE}
}

@article{velickovic,
  title={Graph attention networks},
  author={Velickovic, Petar and Cucurull, Guillem and Casanova, Arantxa and Romero, Adriana and Lio, Pietro and Bengio, Yoshua and others},
  journal={stat},
  volume={1050},
  number={20},
  pages={10--48550},
  year={2017}
}

@article{Chen2025,
title = {Dynamic trend fusion module for traffic flow prediction},
journal = {Applied Soft Computing},
volume = {174},
pages = {112979},
year = {2025},
issn = {1568-4946},
author = {Jing Chen and Haocheng Ye and Zhian Ying and Yuntao Sun and Wenqiang Xu},
}

@inproceedings{Zhou2024,
author = {Zhou, Yicheng and Wang, Pengfei and Dong, Hao and Zhang, Denghui and Yang, Dingqi and Fu, Yanjie and Wang, Pengyang},
title = {Make graph neural networks great again: a generic integration paradigm of topology-free patterns for traffic speed prediction},
year = {2024},
isbn = {978-1-956792-04-1},
booktitle = {Proceedings of the Thirty-Third International Joint Conference on Artificial Intelligence},
articleno = {288},
numpages = {9},
location = {Jeju, Korea},
series = {IJCAI '24}
}

@article{Yang2024,
  title={PSTCGCN: Principal spatio-temporal causal graph convolutional network for traffic flow prediction},
  author={Yang, Shiyu and Wu, Qunyong and Li, Ziwei and Wang, Keyue},
  journal={Neural Computing and Applications},
  pages={1--14},
  year={2024},
  publisher={Springer}
}

@inproceedings{Zhao2023,
    title={Causal Conditional Hidden Markov Model for Multimodal Traffic Prediction},
    author={Zhao, Yu and Deng, Pan and Liu, Junting and Jia, Xiaofeng and Wang, Mulan},
    booktitle={Proceedings of the AAAI Conference on Artificial Intelligence},
    year={2023}
}

@inproceedings{wiegreffe2019,
    title = "Attention is not not Explanation",
    author = "Wiegreffe, Sarah  and
      Pinter, Yuval",
    editor = "Inui, Kentaro  and
      Jiang, Jing  and
      Ng, Vincent  and
      Wan, Xiaojun",
    booktitle = "Proceedings of the 2019 Conference on Empirical Methods in Natural Language Processing and the 9th International Joint Conference on Natural Language Processing (EMNLP-IJCNLP)",
    month = nov,
    year = "2019",
    address = "Hong Kong, China",
    publisher = "Association for Computational Linguistics",
    doi = "10.18653/v1/D19-1002",
    pages = "11--20",

}

@article{lundberg2017,
  title={A unified approach to interpreting model predictions},
  author={Lundberg, Scott M and Lee, Su-In},
  journal={Advances in neural information processing systems},
  volume={30},
  year={2017}
}

@article{bach2015pixel,
  title={On pixel-wise explanations for non-linear classifier decisions by layer-wise relevance propagation},
  author={Bach, Sebastian and Binder, Alexander and Montavon, Gr{\'e}goire and Klauschen, Frederick and M{\"u}ller, Klaus-Robert and Samek, Wojciech},
  journal={PloS one},
  volume={10},
  number={7},
  pages={e0130140},
  year={2015},
  publisher={Public Library of Science San Francisco, CA USA}
}

@inproceedings{zhang2021traffic,
  title={Traffic flow forecasting with spatial-temporal graph diffusion network},
  author={Zhang, Xiyue and Huang, Chao and Xu, Yong and Xia, Lianghao and Dai, Peng and Bo, Liefeng and Zhang, Junbo and Zheng, Yu},
  booktitle={Proceedings of the AAAI conference on artificial intelligence},
  volume={35},
  number={17},
  pages={15008--15015},
  year={2021}
}

@article{murdoch2019,
  title={Definitions, methods, and applications in interpretable machine learning},
  author={Murdoch, W James and Singh, Chandan and Kumbier, Karl and Abbasi-Asl, Reza and Yu, Bin},
  journal={Proceedings of the National Academy of Sciences},
  volume={116},
  number={44},
  pages={22071--22080},
  year={2019},
  publisher={National Academy of Sciences}
}

@inproceedings{franceschi2019,
  title={Learning discrete structures for graph neural networks},
  author={Franceschi, Luca and Niepert, Mathias and Pontil, Massimiliano and He, Xiao},
  booktitle={International conference on machine learning},
  pages={1972--1982},
  year={2019},
  organization={PMLR}
}

@article{jiang2022graph,
  title={Graph neural network for traffic forecasting: A survey},
  author={Jiang, Weiwei and Luo, Jiayun},
  journal={Expert systems with applications},
  volume={207},
  pages={117921},
  year={2022},
  publisher={Elsevier}
}

@inproceedings{jin2020graph,
  title={Graph structure learning for robust graph neural networks},
  author={Jin, Wei and Ma, Yao and Liu, Xiaorui and Tang, Xianfeng and Wang, Suhang and Tang, Jiliang},
  booktitle={Proceedings of the 26th ACM SIGKDD international conference on knowledge discovery \& data mining},
  pages={66--74},
  year={2020}
}

@inproceedings{zhang2023,
  title={Automated spatio-temporal graph contrastive learning},
  author={Zhang, Qianru and Huang, Chao and Xia, Lianghao and Wang, Zheng and Li, Zhonghang and Yiu, Siuming},
  booktitle={Proceedings of the ACM web conference 2023},
  pages={295--305},
  year={2023}
}

@article{he2023,
  title={STGC-GNNs: A GNN-based traffic prediction framework with a spatial--temporal Granger causality graph},
  author={He, Silu and Luo, Qinyao and Du, Ronghua and Zhao, Ling and He, Guangjun and Fu, Han and Li, Haifeng},
  journal={Physica A: Statistical Mechanics and its Applications},
  volume={623},
  pages={128913},
  year={2023},
  publisher={Elsevier}
}

@article{zhu2025,
  title={Traffic Prediction using an Active Causality Recurrent Graph Convolutional Network},
  author={Zhu, Jinde and Yuan, Junhao and Nguyen, Trong-The and Wang, Ruoxi and Zeng, Wu},
  journal={Expert Systems with Applications},
  pages={129506},
  year={2025},
  publisher={Elsevier}
}

% author biography
\vspace{-40pt}
\begin{IEEEbiography}[{\includegraphics[width=1in,height=1.25in,clip,keepaspectratio]{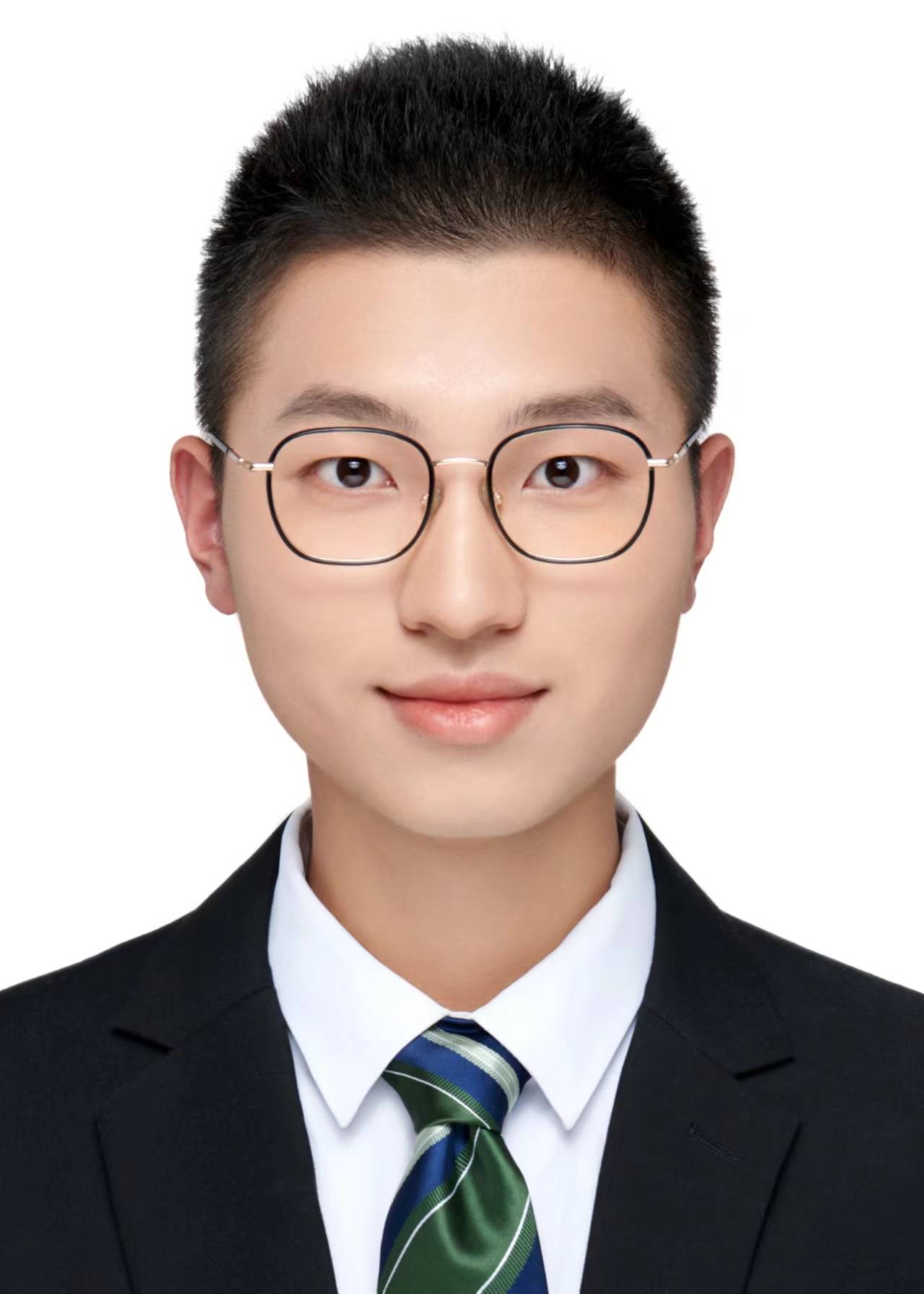}}]{Xuanmian He} is currently pursuing a master's degree in Transportation Engineering at University of California, Berkeley. He received the bachelor's degree in Traffic Engineering from Tongji University in 2025. His research interests include the intersection of artificial intelligence with traffic flow modeling and control, connected and automated vehicles, as well as trustworthy and interpretable decision-making in transportation.
\end{IEEEbiography}

\vspace{-3em}

\begin{IEEEbiography}[{\includegraphics[width=1in,height=1.25in,clip,keepaspectratio]{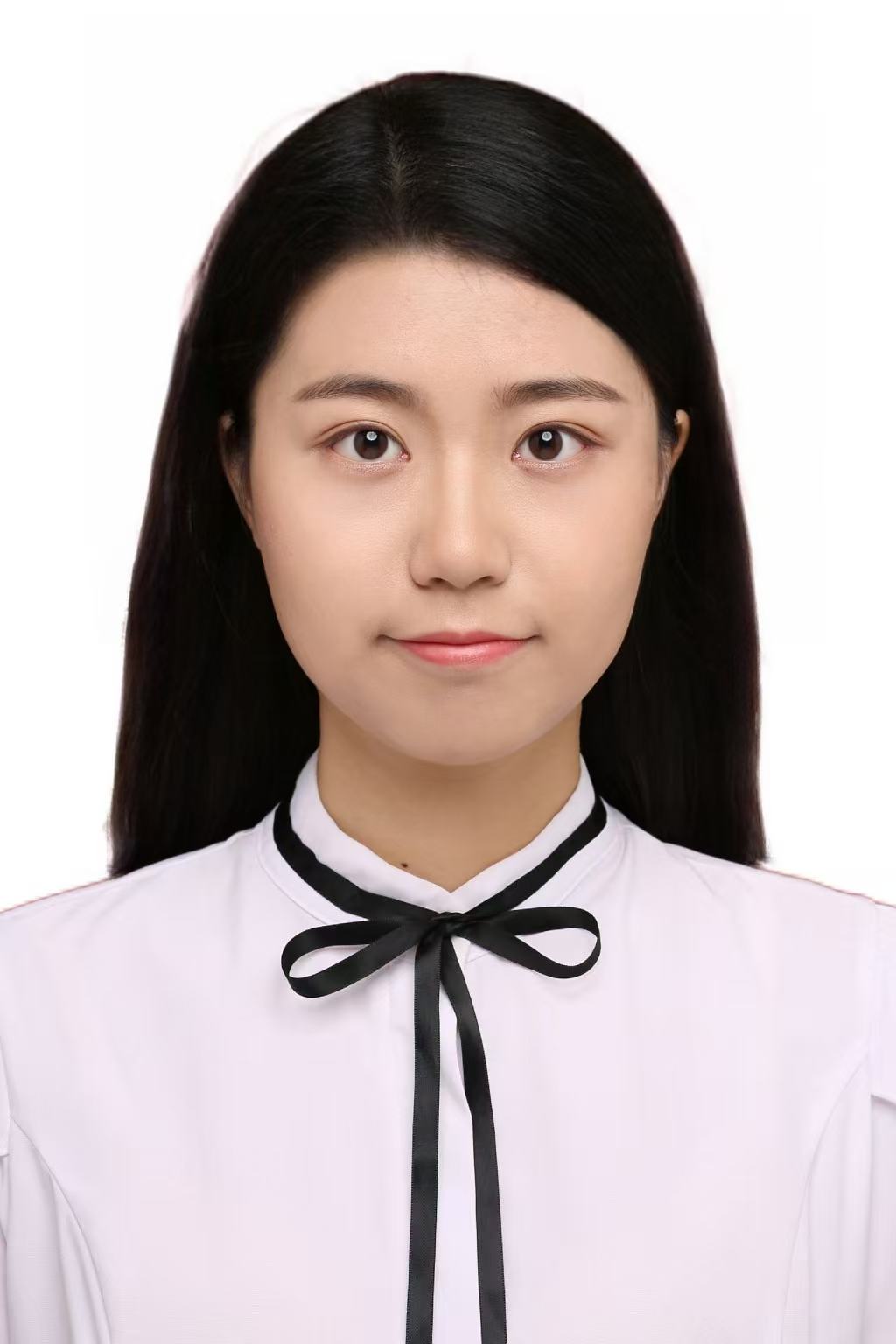}}]{Can Li}
Can Li received the M.S. degree in the Department of Electrical and Computer Engineering from Rutgers University and PhD degree in the School of Computer Science and Engineering from the University of New South Wales. She is currently an Associate Professor in the college of Transportation at Tongji University. Her research focuses on urban big data analytics, deep learning modeling, and transportation foundation models.
\end{IEEEbiography}

\vspace{-3em}

\begin{IEEEbiography}[{\includegraphics[width=1in,height=1.25in,clip,keepaspectratio]{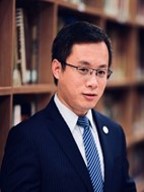}}]{Wanjing Ma} received the B.S. degree in architecture engineering from Chang'an University, Xi'an, China, in 2001, and the M.S. and Ph.D. degrees in traffic information engineering and control from Tongji University, Shanghai, China, in 2004 and 2007, respectively. He is currently a Professor with the College of Transportation, Tongji University. He has published more than 100 articles in Transportation Research Part B: Methodological and other domestic or foreign core journals and academic conferences. His research interests include traffic and control, vehicle-road coordination, and shared mobility.
\end{IEEEbiography}
\end{document}